\documentclass{article}

\usepackage{microtype}
\usepackage{graphicx}
\usepackage{subfigure}
\usepackage{booktabs} 

\usepackage{hyperref}

\usepackage[accepted]{icml2023}

\usepackage{amsmath}
\usepackage{amssymb}
\usepackage{mathtools}
\usepackage{caption}
\usepackage{amsthm}
\usepackage{makecell}
\usepackage{graphicx}
\usepackage{subfigure}
\usepackage{natbib}
\usepackage{bm}
\usepackage{multirow}
\usepackage{multicol}
\usepackage{hyperref}
\usepackage{url}
\usepackage{optidef}
\usepackage{relsize}
\usepackage{xcolor}
\usepackage{booktabs}
\usepackage{algorithm}
\usepackage{algorithmic}
\usepackage{float} 
\usepackage{wrapfig}
\usepackage{bbold}
\usepackage[capitalize,noabbrev]{cleveref}

\theoremstyle{plain}
\newtheorem{theorem}{Theorem}[section]
\newtheorem{proposition}[theorem]{Proposition}
\newtheorem{lemma}[theorem]{Lemma}

\theoremstyle{definition}
\newtheorem{definition}[theorem]{Definition}

\theoremstyle{remark}

\usepackage[textsize=tiny]{todonotes}

\icmltitlerunning{Restricted Orthogonal Gradient Projection for Continual Learning}

\begin{document}

\twocolumn[
\icmltitle{Restricted Orthogonal Gradient Projection for Continual Learning}




\begin{icmlauthorlist}
\icmlauthor{Zeyuan Yang}{xxx}
\icmlauthor{Zonghan Yang}{xxx}
\icmlauthor{Peng Li}{yyy}
\icmlauthor{Yang Liu}{xxx,yyy,zzz,ppp,qqq,rrr}
\end{icmlauthorlist}

\icmlaffiliation{xxx}{Department of Computer Science and Technology, Institute of AI, Tsinghua University}
\icmlaffiliation{yyy}{Institute for AI Industry Research, Tsinghua University}
\icmlaffiliation{zzz}{Beijing National Research Center for Information Science and Technology}
\icmlaffiliation{ppp}{Beijing Academy of Artificial Intelligence}
\icmlaffiliation{qqq}{International Innovation Center of Tsinghua University}
\icmlaffiliation{rrr}{Quan Cheng Laboratory}

\icmlcorrespondingauthor{Peng Li}{lipeng@air.tsinghua.edu.cn}
\icmlcorrespondingauthor{Yang Liu}{liuyang2011@tsinghua.edu.cn}

\icmlkeywords{Machine Learning, Continual Learning, Gradient Projection}

\vskip 0.3in
]



\printAffiliationsAndNotice{\icmlEqualContribution} 

\begin{abstract}
Continual learning aims to avoid catastrophic forgetting and effectively leverage learned experiences to master new knowledge. 
Existing gradient projection approaches impose hard constraints on the optimization space for new tasks to minimize interference, which simultaneously hinders forward knowledge transfer. 
To address this issue, recent methods reuse frozen parameters with a growing network, resulting in high computational costs.
Thus, it remains a challenge whether we can improve forward knowledge transfer for gradient projection approaches \textit{using a fixed network architecture}. 
In this work, we propose the Restricted Orthogonal Gradient prOjection (ROGO) framework. 
The basic idea is to adopt a restricted orthogonal constraint allowing parameters optimized in the direction oblique to the whole frozen space to facilitate forward knowledge transfer while consolidating previous knowledge. 
Our framework requires neither data buffers nor extra parameters.
Extensive experiments have demonstrated the superiority of our framework over several strong baselines.
We also provide theoretical guarantees for our relaxing strategy. 
\end{abstract}

\section{Introduction}

A critical capability for intelligent systems is to continually learn given a sequence of tasks ~\citep{thrun1995lifelong, mccloskey1989catastrophic}. 
Unlike human beings, vanilla neural networks straightforwardly update parameters regarding current data distribution when learning new tasks, suffering from catastrophic forgetting~\citep{mccloskey1989catastrophic, ratcliff1990connectionist,kirkpatrick2017overcoming}. 
Continual learning without forgetting has thus gained increasing attention in recent years~\citep{kurle2019continual, ehret2020continual, ramesh2021model,liu2022continual, tengovercoming}. 
Moreover, an ideal continual learner is expected to not only avoid catastrophic forgetting but also facilitate forward knowledge transfer~\citep{lopez2017gradient}, i.e., leveraging past learning experiences to master new knowledge efficiently and effectively~\citep{parisi2019continual,finn2019online}. 
Without forward knowledge transfer, approaches may have limited performance even with less forgetting~\citep{kemker2018measuring}.

To mitigate forgetting and facilitate forward knowledge transfer, replay-based methods~\citep{lopez2017gradient, shin2017continual, choi2021dual} stores some old samples in the memory, and expansion-based methods~\citep{rusu2016progressive, yoon2017lifelong, yoon2019scalable} expand the model structure to accommodate incoming knowledge.
However, these methods require either extra memory buffers~\citep{parisi2019continual} or a growing network architecture as new tasks continually arrive~\citep{kong2022balancing}, which are always computationally expansive~\citep{de2021continual}.
Thus, promoting performance within a fixed network capacity remains challenging.
Regularization-based methods~\citep{kirkpatrick2017overcoming, de2021continual, aljundi2018memory} penalize the transformation of parameters regarding the corresponding plasticity via regularization terms.
Instead of constraining individual neurons explicitly, recent gradient projection methods~\citep{zeng2019continual, saha2021gradient, wang2021training} further constrain the directions of gradient update, obtaining superior performance.
However, in spite of effectively mitigating forgetting, the limited optimization space also hinders the capability of learning new tasks~\citep{zeng2019continual}, resulting in insufficient forward knowledge transfer.

\begin{figure}[t]
    \centering
    \includegraphics[width=0.95\linewidth]{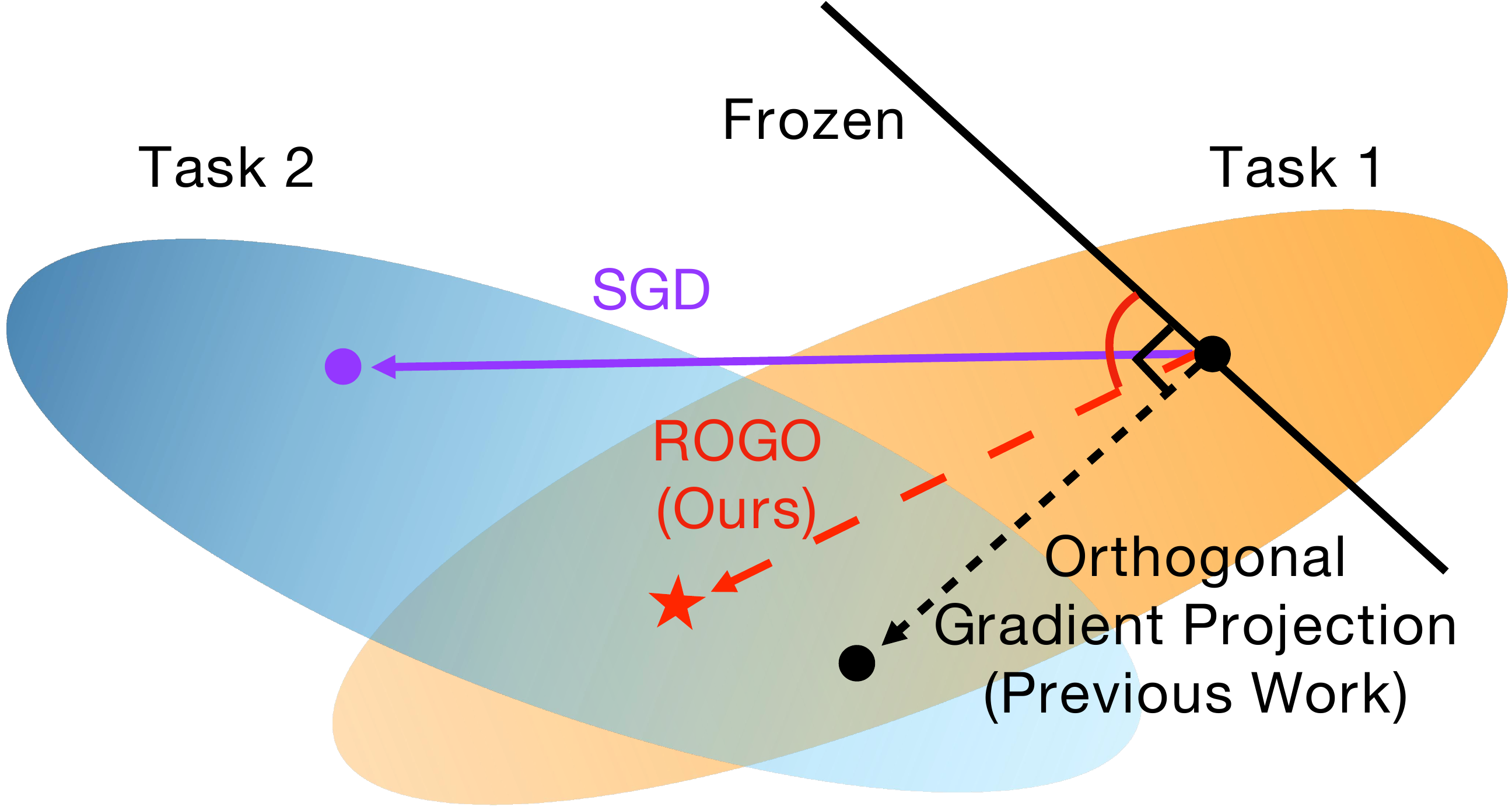}
    \caption{Illustration of our ROGO framework. The lines with arrow denote the gradients on task 2 (pale blue area) after learning task 1 (pale yellow area). We indicated the global optimum by the red star ({\color{red}$\bigstar$}). The black straight line denotes the frozen space, to which gradient projection methods constrain the gradients to be orthogonal.}    
    \label{fig:framework}
    \vspace{-15pt}
\end{figure}

Particularly, we illustrate the optimizing process of traditional gradient projection methods by the black dashed line in Figure~\ref{fig:framework}.
As shown in Figure~\ref{fig:framework}, the strict orthogonal constraint may exclude the global optimum (the red star) from the optimization space.
In other words, constraining the directions of gradient update fails on the plasticity in the stability-plasticity dilemma~\citep{french1997pseudo}.
TRGP~\citep{lin2022trgp} tackles this problem by allocating new parameters within the selected subspace of old tasks as trust regions, suffering a similar computational cost burden as expansion-based methods~\citep{wang2021training}.
Therefore, promoting forward knowledge transfer within a fixed network capacity remains a key challenge for gradient projection methods.

To address this challenge, we propose our Restricted Orthogonal Gradient prOjection (ROGO) framework to facilitate forward knowledge transfer.
Instead of optimizing in an orthogonal direction, our method adopts a restricted orthogonal constraint.
As illustrated by the red line in Figure~\ref{fig:framework}, our framework allows the parameters to be updated in the direction oblique to the original frozen space, which explores a larger optimization space and thus obtains better performance on new tasks.
Specifically, we design a simple yet effective strategy to find the critical subspace, namely the relaxing space, within the frozen space.
The parameters are jointly optimized in the relaxing and original optimization spaces to facilitate forward knowledge transfer.
A complimentary regularization term is introduced to consolidate previous knowledge as well.
Extensive experiments on various continual learning benchmarks demonstrate that our ROGO framework promotes \textbf{forward knowledge transfer} and achieves better classification performance compared with related state-of-the-art approaches.
Moreover, our framework can also be extended as an expansion-based method by storing the parameters of the selected relaxing space, universally surpassing TRGP~\citep{lin2022trgp} and other expansion-based approaches.
We also provide theoretical proof to guarantee the efficiency of our strategy.

\section{Related Work}\label{sec:2}

\subsection{Gradient Projection Methods}
Gradient projection methods constrain the gradients to be orthogonal to the frozen space constructed by previous tasks to overcome forgetting.
OWM~\citep{zeng2019continual} first proposed to modify the gradients upon projector matrices.
OGD~\citep{farajtabar2020orthogonal} further keeps the gradients orthogonal to the space spanned by previous gradients, whereas GPM~\citep{saha2021gradient} computes the frozen space based on old data.
NCL~\citep{kao2021natural} combines the idea of gradient projection and Bayesian weight regularization to mitigate catastrophic forgetting.
In spite of minimizing backward interference, these approaches suffer poor forward knowledge transfer and lack plasticity~\citep{kong2022balancing}.
TRGP~\citep{lin2022trgp} expands the model with trust regions to achieve better performance on new tasks by introducing additional scale parameters.
On the contrary, we focus on facilitating forward knowledge transfer \emph{within a fixed capacity network} by optimizing the parameters in the direction oblique to the frozen space.

\subsection{Regularization-based Methods}
Regularization-based methods introduce regularization terms to the objective function to penalize the modification of parameters, requiring neither data buffers nor extra parameters as gradient projection methods.
EWC~\citep{kirkpatrick2017overcoming} first proposes to constrain the change based on approximated importance weights.
HAT~\citep{serra2018overcoming} learns task-based hard attention to identify important parameters.
Other methods, also called parameter-isolation methods, defy forgetting via freezing the gradient updates of particular parameters~\citep{de2021continual}.
PackNet~\citep{mallya2018packnet} iteratively prunes and allocates parameters subset to incoming tasks, whereas \citet{kumar2021bayesian} facilitate inter-task transfer with the Indian Buffet Process~\citep{griffiths2011indian}.
Instead of restricting individual parameters update, the main idea of our approach is constraining the direction of gradients.

\subsection{Other Methods}
Replay-based methods~\citep{lopez2017gradient, chaudhry2018efficient} maintain a complementary memory for old data, which are replayed during learning new tasks.
Recent approaches~\citep{chenshen2018memory, cong2020gan} deploy auxiliary deep generative models to synthesize pseudo data.
However, including extra data into the current task introduces excessive training time~\citep{de2021continual}.
Expansion-based methods~\citep{yoon2017lifelong, yoon2019scalable, douillard2022dytox} dynamically allocate new parameters or modules to learn new tasks.
While these methods face capacity explosion inevitably after learning a long sequence of tasks.
In contrast, our method maintains a fixed network architecture and requires no previous data. 

\section{Restricted Orthogonal Gradient Projection}
\label{sec:3}

\subsection{Preliminaries}

In a continual learning setting, we consider $T$ tasks arriving as a sequence.
When learning the current task, the datasets of old tasks are inaccessible.
We use an $L$-layer neural network with fixed capacity, and parameters defined as $\mathcal{W} = \{W^l\}_{l=1}^L$, where $W^l$ denotes the parameters in the $l$-th layer.

To mitigate catastrophic forgetting, gradient projection methods construct the representative space, namely the frozen space, based on previous tasks and optimize the parameters orthogonal to the frozen space.
Specifically, for each task $t$, \citet{saha2021gradient} obtain the layer-wise representation space $R^l_t$ by compressing the representation matrix $\mathbf{H}_{t}^l = [h_{t,1}^l, ..., h_{t, N_{t}}^l]$, where $N_t$ denotes the number of samples in task $t$ and $h^l_{t,j}$ denotes the intermediate representation of the $j$-th input.
The constructed representation spaces of previous tasks are then concatenated to be the frozen gradient space $\{ U_{t}^l \}_{l=1}^L$ for future training.
\begin{equation}
    \mathcal{U}_{t} = \{ U_{t}^l \}_{l=1}^L  = \{ R_{1}^l \cup \cdots \cup R_{t}^l \}_{l=1}^L 
    \label{eq:concat-frozen-space}
\end{equation}
During training task $t$, gradients $g_t^l$ are layer-wise constrained to be orthogonal to $U_{t-1}^l$.
Particularly, assuming $\mathbf{B}_{t-1}^l = [u_{t-1,1}^l, ..., u_{t-1, N}^l]$ as the total $N$ orthogonal basis of $U_{t-1}^l$, GPM~\citep{saha2021gradient} modifies the gradients as:
\vspace{-5pt}
\begin{equation}
    \begin{aligned}
    g_t^l&= g_t^l - \mathrm{Proj}_{U_{t-1}^l}(g_{t}^{l}) \\
    &= g_t^l - \mathbf{B}_{t-1}^l ( \mathbf{B}_{t-1}^l )^T g_{t}^{l}
    \end{aligned}
    \label{eq:gpm-train}
\end{equation}
In spite of alleviating forgetting, the strict orthogonal constraint on parameters hinders the forward knowledge transfer by the limited optimization space, thus compromising the performance of new tasks.
TRGP~\citep{lin2022trgp} tackles this problem by selecting old tasks relevant to the current task and expanding corresponding frozen spaces as the trust regions.
The scaled weight projection is further introduced for memory-efficient updating and storing the parameters by scaling the basis.
Considering that task $i$ is selected as the trust region, the scaled weight projection is:
\vspace{-2pt}
\begin{equation}
    \mathrm{Proj}_{U^l_{i}}^{S^l_{i}} (g^l_t) = \mathbf{B}^l_{i} \mathbf{S}^l_i {(\mathbf{B}^l_{i})}^T g^l_t
\end{equation}
where $\mathbf{S}^l_i$ denotes the scale matrix.
For each task $t$, TRGP selects several different tasks as the trust regions $U^l_i$.
During the training phase, gradient $g_t^l$ is modified as:
\vspace{-2pt}
\begin{equation}
g^l_t = g^l_t - \mathrm{Proj}_{U^l_{t-1}} (g^l_t) + \sum_{i} \mathrm{Proj}_{U^l_i}^{S^l_i} (g^l_t)
\label{eq:trgp-train}
\vspace{-2pt}
\end{equation}
The parameters in the trust regions are retrained and the learned scale matrices are stored in the memory.
During the inference phase, TRGP retrieves the model training on each task by replacing the parameters with the scale matrices.
The parameters $W^l_{t,I}$ used for inference on task $t$ are:
\vspace{-2pt}
\begin{equation}
W^l_{t,I} = W^l - \sum_{i} \mathrm{Proj}_{U^l_i} (W^l) + \sum_{i} \mathrm{Proj}_{U^l_i}^{S^l_i} (W^l)
\label{eq:consolidate-parameter}
\end{equation}
However, as tasks come, storing scaling matrices introduces increasing extra parameters.
Our experiments demonstrate that TRGP requires around 5,000\% amount of the parameters regarding the initial network after 20 tasks on MiniImageNet, see Figure~\ref{fig:params}-(b).
Therefore, we propose to facilitate forward knowledge transfer within a fixed network capacity by a restricted orthogonal constraint, 
allowing parameters updated in the direction oblique to the whole frozen space.

\subsection{Overview}

In this section, we introduce our Restricted Orthogonal Gradient prOjection (ROGO) framework.
To better characterize the relationship between the gradient direction and the frozen space, we first define the angle between a given space and a vector in Definition~\ref{def:angle}.

\begin{definition}
(Angle between vector and space) 
We denote the angle between two inputs as $\mathit{\Theta}( \cdot )$ and the inner product between two vectors as $\langle \cdot, \cdot \rangle$.
The angle between a vector $v \in \mathbb{R}^{n}$ and a space $U^{n \times c} \subset \mathbb{R}^{n}$ is defined as the minimum angle between the given vector $v$ and any unit vector $u \in U$:
\vspace{-10pt}
\begin{equation}
    \mathit{\Theta}( v, U ) = \arccos{\underset{u \in U}{\min} \frac{\langle v, u \rangle}{\Vert v \Vert}} 
\end{equation}
\label{def:angle}
\vspace{-15pt}
\end{definition}
Instead of ${90}^{\circ}$ in traditional gradient projection methods, we aim to relax the angle between the frozen space and the modified gradient to be an acute angle.
However, directly rotating a vector in a high-dimensional space is too flexible to control.
Hence, we notice the Proposition~\ref{pro:oblique}.
\begin{proposition}
    Given the full space $R$ and a random space $U \subset R$, for any vector $v \in U^{\perp}$, where $U^{\perp} = R \backslash U$ denotes the orthogonal complement of $U$, it can be modified to be oblique to the given space $U$ in any angle, by scaling and combining a vector $v'$ within a selected subspace $U' \subseteq U$.
    \label{pro:oblique}
\end{proposition}
\vspace{-2pt}
According to Proposition~\ref{pro:oblique}, we can manipulate the gradient direction to be oblique to the entire frozen space $U^l_{t-1}$ in any target angle by modifying the parameters within an appropriate subspace of $U^l_{t-1}$.
Therefore, for each task $t$, instead of directly rotating the gradient, we introduce the layer-wise relaxing space $V_t^l \subseteq U^l_{t-1}$.
During the training phase, we modify the gradients as:
\vspace{-2pt}
\begin{equation}
    g^l_t = g^l_t - \mathrm{Proj}_{U^l_{t-1} \backslash V^l_t}(g^l_t)
\end{equation}
By jointly optimizing the parameters within the selected relaxing space $V_t^l$ and the original optimization space, the strict orthogonal constraint is relaxed.
Hence, with a null relaxing space, namely $\mathrm{dim}(V^l_t) = 0$, our framework works exactly the same as GPM, under the orthogonal constraint.
Therefore, our framework can be viewed as a generalization of GPM by regulating the relaxing space $V^l_t$.
The relaxing space selection thus becomes the key procedure.

In Section~\ref{sec:3-3}, we introduce our searching strategy for the relaxing space.
Moreover, given the relaxing space $V_t^l$, we involve complementary regularization loss to better consolidate previous knowledge.
The detailed procedure is provided in Section~\ref{sec:3-4}.
For better illustration, we provide the procedure of our ROGO framework in Algorithm~\ref{alg:main}.
Besides, we propose ROGO-Exp in Section~\ref{sec:3-5}, involving additional parameters as expansion-based methods, to further validate the flexibility and effectiveness of our framework.

\begin{algorithm}
    \renewcommand{\algorithmicrequire}{\textbf{Input:}}
	\renewcommand{\algorithmicensure}{\textbf{Output:}}
	\caption{Restricted orthogonal gradient projection}
	\label{alg:main}
	\begin{algorithmic}[1]
		\STATE Initiate the frozen spaces $\mathcal{U}_0=\{U^l_0\}^L_{l=1}$ as $\varnothing$s and optimize $\mathcal{W}_1$ for task 1.
		\STATE Compute frozen space $\mathcal{U}_1$ with representation matrices.
        \FOR {$t \in 2,...,T$}
            \STATE Initiate the relaxing space $\{V^l_{t}\}_{l=1}^L$ as $\varnothing$s.
    		\REPEAT
		    \STATE Fine-tune for predefined $e_t$ epochs.
		    \STATE Determine additional relaxing space $\{V^{l,new}_t\}_{l=1}^L$ by the searching strategy in Section~\ref{sec:3-3}.
		    \STATE $V^l_t \leftarrow V^l_t \cup V^{l,new}_t$.
            \UNTIL $V^{l,new}_t$ is $\varnothing$.
		    \STATE Optimize $\{W^l_t\}_{l=1}^L$ within the space orthogonal to $U_{t-1}^l \backslash V^l_t$ with the regularization terms in Section~\ref{sec:3-4}.
            \STATE Determine the representation space $\{R_t^l\}_{l=1}^L$.
            \STATE Update the frozen space by $U_t^l = U_{t-1}^l \cup R_t^l$.
        \ENDFOR
	\end{algorithmic}
\end{algorithm}

\subsection{Searching the Relaxing Space}
\label{sec:3-3}

To determine the relaxing space $V^l_t$, we first construct the representation space $R_{g,t}^l$ by the top-$k$ principal eigenvectors of the representative matrix $[g_{t,1}^l, ..., g_{t, N_{t}}^l]$, where $g^l_{t,j}$ denotes the gradient of the $j$-th input.
The estimated importance is further characterized by the angle from $R_{g,t}^l$.
Here we define that a vector $d$ is \emph{relaxable} when:
\begin{equation}
    \mathit{\Theta}( d, R_{g,t}^l ) \leq \gamma^l_t
    \label{eq:angle}
\end{equation}
where $\gamma^l_t$ is a predefined threshold.
For task $t$, we aim to find the relaxing space $V^l_{t} \subseteq U^l_{t-1}$ spanned all by \emph{relaxable} vectors in $U^l_{t-1}$, namely:
\begin{equation}
    \left\{\begin{array}{ll}
        \underset{\ \, u \in V^l_t \ \,}{\max} \mathit{\Theta} ( u, R_{g,t}^l ) \leq \gamma^l_t &  \\
        \underset{v \in U^{l,c}_{t-1}}{\min} \mathit{\Theta} ( v, R_{g,t}^l ) > \gamma^l_t & 
    \end{array}\right.
    \label{eq:criteria}
\end{equation}
where $U^{l,c}_{t-1} = U^l_{t-1} \backslash V^l_t$ denotes the complemented subspace of $V^l_t$ with respect to $U^l_{t-1}$.
Criterion~(\ref{eq:criteria}) guarantees that all vectors in $V^l_t$ are \emph{relaxable} and there remains none \emph{relaxable} vector in $U^{l,c}_{t-1}$.
However, it is hard to construct the appropriate $V_t^l$ directly from the high-dimensional $U_{t-1}^l$.

Therefore, we propose a simple yet efficient strategy to find $V^l_t$.
Initiating $V^l_t$ as $\varnothing$, we iteratively select the closest vector to $R_{g,t}^l$ within $U^{l,c}_{t-1}$ by ${\mathop{\arg\min}} \mathit{\Theta} ( d, R_{g,t}^l )$ and then append it into $V_t^l$ as the basis if it is \emph{relaxable}.
This procedure is repeated until no \emph{relaxable} vector is left, thus obtaining the target $V^l_t$ consisting of all \emph{relaxable} vectors within $U^l_{t-1}$.
The pseudo-code of our searching strategy is provided in Algorithm~\ref{alg:RSS} in Appendix~\ref{app:algo}.
\begin{theorem}
Denote $\mathcal{S}=\{u \vert \mathit{\Theta} ( u, R^l_{g,t} ) \leq \gamma^l_t \ \mathrm{and} \ u \in U^l_{t-1}\}$ as the whole solution set, the obtained relaxing space $V^l_t$ takes up the maximum space in $\mathcal{S}$.
\label{the:max-rank}
\end{theorem}
To further substantiate the effectiveness of our searching strategy, here we introduce Theorem~\ref{the:max-rank}.
According to Theorem~\ref{the:max-rank}, our strategy guarantees to find the maximum space within the whole solution set satisfying criterion~(\ref{eq:criteria}).
The detailed proof is provided in Appendix~\ref{app:proof-max-rank}.

\begin{theorem}
Denote $k_r$ as the dimension of the representation space $R^l_{g,t}$ and $k_v$ as the dimension of the relaxing space $V^l_t$, $k_v \leq k_r$, regardless of the frozen space $U^l_{t-1}$. 
\label{the:rank-k}
\end{theorem}

Moreover, we provide theoretical analysis on the upper bound of the dimension of $V^l_t$, which is also the number of iterations, to investigate the efficiency of our strategy.
Here we introduce Theorem~\ref{the:rank-k}, which guarantees that the dimension of $V^l_t$ is no more than of the representation space $R^l_{g,t}$.
In other words, the number of iterations is bounded.
Hence, as $R^l_{g,t}$ is constructed by the top-$k$ principal eigenvectors, the dimension of $V^l_t$ is further regulated by hyperparameters, of which the ablation study is provided in Section~\ref{sec:5}.
We also include the detailed proof in Appendix~\ref{app:proof-rank-k}.

\subsection{Constrained Update in the Relaxing Space}
\label{sec:3-4}

Given the selected relaxing space $V^l_t$, the optimization space is enlarged, thus facilitating forward knowledge transfer.
In the meanwhile, we also want to consolidate previous knowledge stored within $V^l_t$.
One direct way is to fine-tune the parameters with additional regularization terms such as EWC~\citep{kirkpatrick2017overcoming}.
However, regularization terms are designed for explicit parameters, which are not applicable to implicit subspace in our framework.
Therefore, we borrow the scaled weight projection~\citep{lin2022trgp} to modify explicit parameters instead.
With the scale matrix $\mathbf{S}^l_t$, the gradient $g^l_t$ is modified as:
\vspace{-5pt}
\begin{equation}
    g^l_t = g^l_t - \mathrm{Proj}_{U^l_{t-1}} (g^l_t) + \mathrm{Proj}_{V^l_t}^{S^l_t} (g^l_t)
    \label{eq:rogo-train}
    \vspace{-5pt}
\end{equation}
Notice that we update the parameters within $\mathcal{V}^l_t$ with $\mathbf{S}^l_t$ after training each task, instead of storing the parameters for inference as TRGP. 
Parameters within $\mathcal{V}^l_t$ are then restrictedly fine-tuned by adding regularization terms on $\mathbf{S}^l_t$ of instead of directly on parameters.
Specifically, the objective function of task $t$ is:
\vspace{-5pt}
\begin{equation}
    \mathcal{L}_t = \mathcal{L}(\mathcal{W}_t, \mathcal{D}^{(t)}) + \sum_{l=1}^L \beta_l \Vert \mathbf{S}^l_t - \mathbb{1} (\mathbf{S}^l_t) \Vert^2_2
    \label{eq:loss-with-re}
    \vspace{-5pt}
\end{equation}
where $\mathbb{1}(\cdot)$ denotes the identity matrix with the same shape as the input matrix and $\beta_l$ is the weight of the regularization term for layer $l$.
During backpropagation, gradients within the relaxing space $V^l_t$ are regulated by $\mathbf{S}^l_t$.
To further validate our strategy, we also equip TRGP with the above regularization terms for better comparison in Section~\ref{sec:4-3}.
Generally, in our framework, we adopt our searching strategy to determine the relaxing space and modify those parameters with constraints on the scaling matrices.

However, during training, the direction of gradients shifts sharply and frequently due to the steep learning scope of deep neural networks.
Diverse subspaces would be selected in different training phases.
Therefore, we iteratively examine whether there remains \emph{relaxable} vectors within the remaining frozen space $U^{l,c}_{t-1}$ after limited epochs and then search for additional relaxing space $V^{l, new}_t$.
If additional relaxing space $V^{l, new}_t$ is involved, we expand the scaling matrix with identity matrices to accommodate the increased space.
On the contrary, TRGP maintains a fixed-size scaling matrix throughout training.
As the $U^l_{t-1}$ is fixed for each task, the number of iterations of our strategy is naturally limited.
In the implementation, we further constrain the maximum number of iterations for efficiency.

In general, by iteratively searching for the relaxing space and modifying those gradients with additional regularization loss, our restricted orthogonal method optimizes the parameters in the direction oblique to the whole frozen space, thus facilitating forward knowledge transfer while consolidating previous knowledge within a fixed network capacity.

\subsection{Extensions}
\label{sec:3-5}

To further validate our framework, we propose ROGO-Exp, a modified version of our proposed ROGO, directly storing the parameters within the relaxing space.
Similar to TRGP, for each task $t$, we retrieve the corresponding relaxing space $\{V^l_t\}^L_{l=1}$ and the scale matrices $\{\mathbf{S}^l_t\}^L_{l=1}$ during the inference phase.
The modified parameters $W_{t,I}^l$ used for inference on task $t$ are:
\vspace{-5pt}
\begin{equation}
    W_{t,I}^l = W^l - \mathrm{Proj}_{V_t^l} (W^l) + \mathrm{Proj}_{V^l_t}^{S^l_t} (W^l)
\vspace{-5pt}
\end{equation}
where $W^l$ denotes the parameters of layer $l$ of current network.
By replacing the parameters in the relaxing space with the parameters optimized in task $t$, the model achieves better performance, at the price of expansive computational cost in long task sequences.
Further experimental results validate the efficiency of our ROGO-Exp against state-of-the-art expansion-based methods. 

\section{Experiments}
\label{sec:4}

\begin{table*}[t]
\caption{Comparison of average final accuracy ACC and forward knowledge transfer $\Omega_{new}$. $^{\star}$ denotes under the non-incremental setting and $^{\dagger}$ denotes requiring an extra data buffer. For each task, we mark the best and the second best performance in \textbf{bold} and \underline{underline} respectively. All results reported are averaged over 5 runs.}
\vspace{-5pt}
\label{tab:main-results}
\begin{center}
\scalebox{0.82}{
\begin{tabular}{lrrrrrrrr}
\toprule
\multirow{2}{*}{\bf Method}  & \multicolumn{2}{c}{\bf CIFAR-100 Split} & \multicolumn{2}{c}{\bf MiniImageNet} & \multicolumn{2}{c}{\bf PMNIST} & \multicolumn{2}{c}{\bf Mixture} \\
\cmidrule(lr){2-3} \cmidrule(lr){4-5} \cmidrule(lr){6-7} \cmidrule(lr){8-9}
& ACC (\%) & $\Omega_{new}$ (\%) & ACC (\%) & $\Omega_{new}$ (\%) & ACC (\%) & $\Omega_{new}$ (\%) & ACC (\%) & $\Omega_{new}$ (\%) \\
\midrule
Multitask$^{\star}$ & 79.58 $\pm$ 0.54 & - & 69.46 $\pm$ 0.62 & - & 96.70 $\pm$ 0.02 & - & 81.29 $\pm$ 0.23 & - \\
\midrule
A-GEM$^{\dagger}$ & 63.98 $\pm$ 1.22 & 77.48 $\pm$ 0.40 & 57.24 $\pm$ 0.72 & 67.55 $\pm$ 1.20 & 83.56 $\pm$ 0.16 & 97.42 $\pm$ 0.03 & 59.86 $\pm$ 1.01 & 85.87 $\pm$ 0.42 \\
ER\_Res$^{\dagger}$ & 71.73 $\pm$ 0.63 & 77.13 $\pm$ 0.18 & 58.94 $\pm$ 0.85 & 69.48 $\pm$ 0.42 & 87.24 $\pm$ 0.53 & 97.37 $\pm$ 0.05 & 75.07 $\pm$ 0.55 & 86.35 $\pm$ 0.22 \\
\midrule
EWC & 68.80 $\pm$ 0.88 & 69.87 $\pm$ 1.09 & 52.01 $\pm$ 2.53 & \underline{63.45 $\pm$ 2.80} & 89.97 $\pm$ 0.57 & 93.13 $\pm$ 0.70 & 69.62 $\pm$ 2.69 & 74.41 $\pm$ 1.00 \\
HAT & 72.06 $\pm$ 0.50 & 71.50 $\pm$ 0.62 & 59.78 $\pm$ 0.57 & 62.63 $\pm$ 0.63 & - & - & \underline{77.54 $\pm$ 0.18} & 79.26 $\pm$ 0.15 \\
OGD & 70.96 $\pm$ 0.32 & \underline{73.86 $\pm$ 0.57} & 59.83 $\pm$ 0.62 & 62.02 $\pm$ 1.24 & 82.56 $\pm$ 0.66 & \bf 97.40 $\pm$ 0.06 & 74.38 $\pm$ 0.27 & 81.97 $\pm$ 0.53 \\
GPM & \underline{72.48 $\pm$ 0.40} & 71.53 $\pm$ 0.54 & \underline{60.41 $\pm$ 0.61} & 61.63 $\pm$ 0.60 & \underline{93.91 $\pm$ 0.16} & \underline{96.56 $\pm$ 0.04} & 77.49 $\pm$ 0.68 & \underline{82.13 $\pm$ 0.37} \\
ROGO & \bf 74.04 $\pm$ 0.35 & \bf 75.22 $\pm$ 0.36 & \bf 63.66 $\pm$ 1.24 & \bf 63.81 $\pm$ 0.39 & \bf 94.20 $\pm$ 0.11 & 96.26 $\pm$ 0.10 & \bf 77.91 $\pm$ 0.45 & \bf 82.21 $\pm$ 0.42 \\
\bottomrule
\end{tabular}
}
\end{center}
\vspace{-8pt}
\end{table*}

\begin{figure*}[t]
    \centering
    \includegraphics[width=\linewidth]{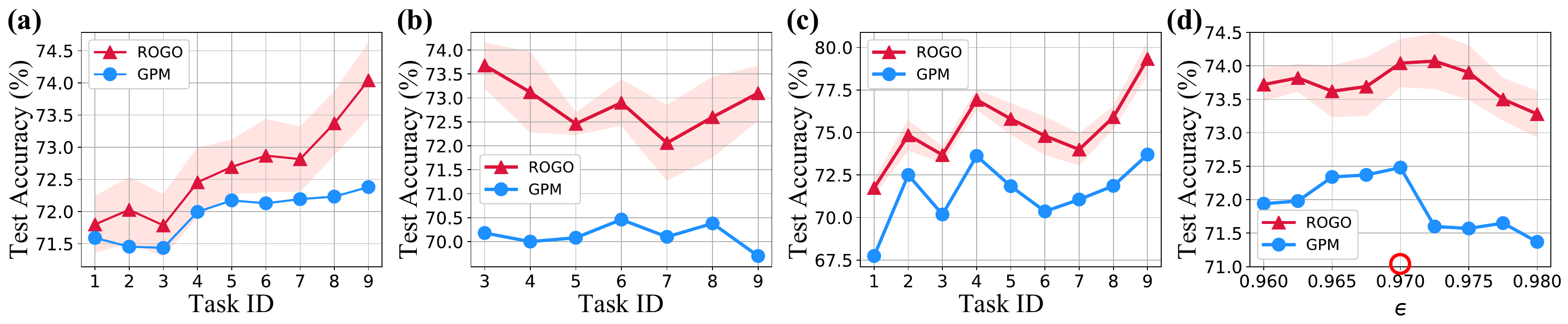}
    \caption{Results on CIFAR-100 Split setting: (a) average accuracy after learning each task; (b) accuracy evolution of a randomly selected task; (c) accuracy tested on task $i$ after learning task $i$; (d) average accuracy of different $\epsilon$. The optimum $\epsilon$ value of GPM is annotated by a red circle ``\textcolor{red}{o}'', and the shaded area indicates the standard deviation. }
    \label{fig:res-cifar}
    \vspace{-10pt}
\end{figure*}

\subsection{Experimental Setup}
\label{sec:4-1}

\textbf{Datasets:}
Following \citet{saha2021gradient}, we evaluate our framework on CIFAR-100 Split~\citep{krizhevsky2009learning}, MiniImageNet~\citep{vinyals2016matching}, Permuted MNIST (PMNIST)~\citep{kirkpatrick2017overcoming}.
Moreover, we introduce the Mixture benchmark, first proposed by~\citep{serra2018overcoming}, consisting of eight datasets.
In this work, we conduct experiments on Mixture with the seven tasks as a sequence except for TrafficSigns\footnote{We fail to obtain TrafficSigns, as all the links provided in~\citep{stallkamp2011german, serra2018overcoming, saha2021gradient} are expired.}.
Details and statistics of the datasets can be found in Appendix~\ref{app:dataset-stat}.
Moreover, we include the details of network architectures in Appendix~\ref{app:model-detail}.

\textbf{Baselines:} We compare ROGO with competitive and well-established methods within a fixed network capacity.
For regularization-based methods, we compare against EWC~\citep{kirkpatrick2017overcoming} and HAT~\citep{serra2018overcoming}.
For gradient projection methods, we consider OGD~\citep{farajtabar2020orthogonal} and GPM~\citep{saha2021gradient}.
For replay-based methods, we adopt ER\_Res~\citep{chaudhry2019continual} and A-GEM~\citep{chaudhry2018efficient}. The memory buffer size for PMNIST, CIFAR-100 Split, MiniImageNet, and Mixture are 1,000, 2,000, 500 and 3,000, respectively.
Moreover, we consider TRGP~\citep{lin2022trgp} as a competitive approach under an expansion setting.
Implementation details are listed in Appendix~\ref{app:implementation-detail}.

\textbf{Metrics:}

Denote $A_{i,j}$ as the test accuracy of task $j$ after learning task $i$ and $b_i$ as the test accuracy of task $i$ at random initialization.
We employ the following three evaluation metrics.

\vspace{-5pt}
\begin{itemize}
    \item Average Accuracy (ACC)~\citep{mirzadeh2020understanding}: the average test accuracy evaluated after learning all tasks, defined as $\frac{1}{T} \sum^T_{i=1} A_{T,i}$.
    \item Backward Transfer (BWT)~\citep{lopez2017gradient}: the average accuracy decrease after learning all tasks, defined as $\frac{1}{T-1} \sum^{T-1}_{i=1} (A_{T,i} - A_{i,i})$.
    \item Forward Transfer ($\Omega_{new}$)~\citep{kemker2018measuring}: the average accuracy of new tasks, defined as $\frac{1}{T-1} \sum^T_{i=2} (A_{i,i}-b_i)$. As $b_i$ stays still across different approaches, we consider $\frac{1}{T-1} \sum^T_{i=2} A_{i,i}$.
\end{itemize}
\vspace{-5pt}

Besides, FWT~\citep{lopez2017gradient} reflects the influence of the observed tasks on new tasks in a zero-shot manner, defined as $\frac{1}{T-1} \sum^T_{i=2} (A_{i-1,i}-b_i)$.
In this paper, we mainly focus on $\Omega_{new}$, while FWT results are also provided.
Detailed definitions are provided in Appendix~\ref{app:metric}.

\subsection{Main Results}
\label{sec:4-2}

We show the comparative results on four benchmarks in Table~\ref{tab:main-results}.
The accuracy results of major baselines are adopted from GPM~\citep{saha2021gradient} and we implement all the baselines to get the forward knowledge transfer and forgetting results.
Implementation details are provided in Appendix~\ref{app:implementation-detail}.
We run each experiment five times and report the mean results and the standard deviation.
Other results including the forgetting are provided in Appendix~\ref{app:bwt}.

According to Table~\ref{tab:main-results}, ROGO dominates EWC across all benchmarks and generally outperforms HAT.
Although HAT obtains comparable accuracy on Mixture, ROGO gains 2.7\% better $\Omega_{new}$ on average.
For gradient projection methods, compared with GPM, ROGO achieves over 1\% and 3\% higher ACC on CIFAR-100 Split and MiniImageNet respectively while improving the forward knowledge transfer.
We observe that OGD+ achieves the better $\Omega_{new}$ on several datasets.
Whereas, it fails on the accuracy with significant forgetting.
Also, we notice that A-GEM and ER\_Res gain better $\Omega_{new}$, in spite of the inferior accuracy compared with GPM and especially ROGO.
We assume that this is because replay-based methods explore the full optimization space with extra data, while both regularization-based and gradient projection methods constrain the optimization space to mitigate forgetting.
Other results including BWT and FWT are provided in Appendix~\ref{app:fwt}.
In general, ROGO obtains the best accuracy and improves forward knowledge transfer without extra data buffers across all datasets.

\begin{table*}[t]
    \centering
    \caption{Comparison of average accuracy and forward knowledge transfer with TRGP under an expansion setting. The percentages indicate the ratios of the rank of the relaxing space with respect to the frozen space. For each task, we mark the best and the second best performance in \textbf{bold} and \underline{underline}. Detailed results are provided in Appendix~\ref{app:others}.}
    \label{tab:trgp-compare-exp}
    \scalebox{0.8}{
    \begin{tabular}{lcccccccc}
        \toprule
        \multirow{3}{*}{Methods} & \multicolumn{2}{c}{\multirow{2}{*}{TRGP}} & \multicolumn{6}{c}{ROGO-Exp} \\
         & & & \multicolumn{2}{c}{50\%} & \multicolumn{2}{c}{80\%} & \multicolumn{2}{c}{T\%} \\
        \cmidrule(lr){2-3} \cmidrule(lr){4-5} \cmidrule(lr){6-7} \cmidrule(lr){8-9}
        & ACC (\%) & $\Omega_{new}$ (\%) & ACC (\%) & $\Omega_{new}$ (\%) & ACC (\%) & $\Omega_{new}$ (\%) & ACC (\%) & $\Omega_{new}$ (\%) \\
        \midrule
        CIFAR & 74.46 $\pm$ 0.22 & 75.01 $\pm$ 0.22 & 74.90 $\pm$ 0.37 & \underline{75.68 $\pm$ 0.43} & \underline{74.97 $\pm$ 0.30} & 75.46 $\pm$ 0.17 & \bf 75.34 $\pm$ 0.61 & \bf 75.86 $\pm$ 0.42 \\
        PMNIST & 96.34 $\pm$ 0.11 & 97.07 $\pm$ 0.14  & 96.44 $\pm$ 0.20 & 97.07 $\pm$ 0.10 & \underline{96.76 $\pm$ 0.15} & \underline{97.20 $\pm$ 0.08} & \bf 97.01 $\pm$ 0.08 & \bf 97.26 $\pm$ 0.05 \\
        MiniImageNet & 61.78 $\pm$ 1.94 & \bf 63.08 $\pm$ 1.40 & \bf 63.46 $\pm$ 0.85 & \underline{62.76 $\pm$ 0.64} & 62.57 $\pm$ 1.32 & 62.32 $\pm$ 1.19 & \underline{62.78 $\pm$ 1.00} & 62.42 $\pm$ 1.28 \\
        Mixture & \underline{83.54 $\pm$ 1.15} & \bf 84.88 $\pm$ 0.95 & 82.45 $\pm$ 0.49 & 84.13 $\pm$ 0.25 & 83.33 $\pm$ 0.31 & \underline{84.60 $\pm$ 0.20} & \bf 83.62 $\pm$ 0.26 & 84.44 $\pm$ 0.42 \\
        \bottomrule
    \end{tabular}
    }
\end{table*}

\begin{table*}[t]
    \centering
    \caption{Comparison of average accuracy and forward knowledge transfer with TRGP within a fixed network capacity. We modify TRGP as TRGP-Reg with similar regularization terms. $\beta$ indicates the regularization weight. For each task, we mark the best and the second best performance in \textbf{bold} and \underline{underline}. Detailed results are provided in Appendix~\ref{app:others}.}
    \label{tab:trgp-compare-non-exp}
    \scalebox{0.8}{
    \begin{tabular}{lcccccccc}
        \toprule
        \multirow{3}{*}{Methods} & \multicolumn{6}{c}{TRGP-Reg} & \multicolumn{2}{c}{\multirow{2}{*}{ROGO}} \\
         & \multicolumn{2}{c}{$\beta=1$} & \multicolumn{2}{c}{$\beta=5$} & \multicolumn{2}{c}{$\beta=50$} & & \\
        \cmidrule(lr){2-3} \cmidrule(lr){4-5} \cmidrule(lr){6-7} \cmidrule(lr){8-9}
        & ACC (\%) & $\Omega_{new}$ (\%) & ACC (\%) & $\Omega_{new}$ (\%) & ACC (\%) & $\Omega_{new}$ (\%) & ACC (\%) & $\Omega_{new}$ (\%) \\
        \midrule
        CIFAR & 72.01 $\pm$ 0.30 & \underline{73.17 $\pm$ 0.24} & 71.96 $\pm$ 0.40 & 72.64 $\pm$ 0.67 & \underline{72.49 $\pm$ 0.09} & 72.67 $\pm$ 0.26 & \bf 74.04 $\pm$ 0.35 & \bf 75.22 $\pm$ 0.36 \\
        PMNIST & \underline{76.57 $\pm$ 2.69} & 94.95 $\pm$ 0.23 & 75.50 $\pm$ 3.96 & 95.47 $\pm$ 0.30 & 76.56 $
        \pm$ 3.83 & \underline{95.88 $\pm$ 0.19} & \bf 94.20 $\pm$ 0.11 & \bf 96.26 $\pm$ 0.10 \\
        MiniImageNet & 55.81 $\pm$ 2.43 & 59.14 $\pm$ 1.00 & \underline{58.81 $\pm$ 2.24} & \underline{62.05 $\pm$ 0.74} & 22.69 $\pm$ 0.39 & 20.39 $\pm$ 0.36 & \bf 63.66 $\pm$ 1.24 & \bf 63.81 $\pm$ 0.39 \\
        Mixture & 73.31 $\pm$ 1.05 & \bf 83.64 $\pm$ 0.38 & \underline{74.71 $\pm$ 0.85} & \underline{83.27 $\pm$ 0.24} & 17.36 $\pm$ 0.86 & 7.27 $\pm$ 0.98 & \bf 77.91 $\pm$ 0.45 & 82.21 $\pm$ 0.42 \\
        \bottomrule
    \end{tabular}
    }
    \vspace{-5pt}
\end{table*}

\begin{table*}[t]
\centering  
\caption{Ablation study of $\zeta = \cos{\gamma}$ and $\beta$ on CIFAR100-Split, where $\gamma$ is the threshold for the relaxing strategy and $\beta$ is the regularization weight. $\zeta_{conv}$ and $\zeta_{fc}$ denote the threshold for convolutional and fully connected layers, respectively.}  
\vspace{-5pt}
\subtable[Ablation study of $\zeta$. For each $\zeta$ combination, we report the relaxing ratio of each layer.]{  
        \scalebox{0.9}{
        \begin{tabular}{rr|rrrrr|rr}
        \toprule
        $\zeta_{cong}$ & $\zeta_{fc}$ & \multicolumn{1}{r}{Conv1} & \multicolumn{1}{r}{Conv2} & \multicolumn{1}{r}{Conv3} & \multicolumn{1}{r}{Fc1} & \multicolumn{1}{r}{Fc2} & \multicolumn{1}{c}{ACC} & \multicolumn{1}{c}{BWT}  \\
        \midrule
        0.20 & 0.20 & 38.09\% & 34.63\% & 51.11\% & 80.75\% & 60.56\% & 71.97\% & -2.97\% \\
        0.50 & 0.50 & 30.26\% & 28.75\% & 40.53\% & 34.78\% & 11.26\% & 72.58\% & -2.41\% \\
        0.80 & 0.80 & 28.51\% & 18.71\% & 25.68\% & 5.94\% & 0.23\% & 73.34\% & -2.04\% \\
        0.90 & 0.90 & 27.87\% & 14.16\% & 18.65\% & 1.08\% & 0.07\% & 73.68\% & -1.76\% \\
        \bf 0.95 & \bf 0.90 & 26.31\% & 10.51\% & 13.63\% & 0.97\% & 0.04\% & \bf 74.04\% & -1.43\% \\
        0.95 & 0.95 & 24.12\% & 10.52\% & 13.83\% & 0.08\% & 0.00\% & 73.87\% & -1.53\% \\
        0.99 & 0.99 & 18.66\% & 5.51\% & 9.51\% & 0.00\% & 0.00\% & 73.70\% & \bf -1.34\% \\
        \bottomrule
        \end{tabular}
        }
       \label{tab:ablation-zeta}  
}  
\quad
\subtable[Ablation study of $\beta$.]{   
        \scalebox{0.9}{
        \begin{tabular}{r|rr}
        \toprule
        $\beta$ & ACC & BWT \\
        \midrule
        0.0 & 72.43\% & -3.03\% \\
        0.1 & 72.94\% & -2.57\% \\
        0.5 & 73.68\% & -1.88\% \\
        \bf 1.0 & \bf 74.04\% & -1.34\% \\
        5.0 & 73.77\% & -1.42\% \\
        10.0 & 73.42\% & -0.87\% \\
        50.0 & 72.65\% & \bf -0.83\% \\
        \bottomrule
        \end{tabular}
        }
       \label{tab:ablation-beta}  
}  
\label{tab:hyper-param}
\vspace{-10pt}
\end{table*} 

Moreover, we compare the average accuracy after learning each task on CIFAR-100 Split with GPM, as GPM achieves the highest accuracy among selected baselines.
As shown in Figure~\ref{fig:res-cifar}-(a), ROGO increasingly gains superior performance by a larger margin, as the optimization space of GPM is gradually constrained by accumulated frozen spaces.
We provide detailed results on other benchmarks in Appendix~\ref{app:final-acc}.
For better illustration, we also exhibit the accuracy evolution of specific tasks during sequential training in Figure~\ref{fig:res-cifar}-(b).
Here we choose the second task on CIFAR-100 Split.
According to the results, ROGO universally outperforms GPM over the task sequence.
Results of three randomly selected tasks are provided in Appendix~\ref{app:acc-evo} respectively.
In addition, we observe the accuracy on new tasks in Figure~\ref{fig:res-cifar}-(c).
Similar to the average accuracy, ROGO achieves increasingly better ${\Omega}_{new}$ by relaxing the strict orthogonal constraint.
Results on other benchmarks included in Appendix~\ref{app:fwt} further substantiate this phenomenon.

To understand our relaxing strategy better, we further conduct experiments on different thresholds $\epsilon$, which regulate the dimension of the frozen space.
\citet{saha2021gradient} argue that $\epsilon$ mediates the stability-plasticity dilemma by controlling the frozen space and thus is critical for GPM.
However, ROGO enables the frozen space to be adaptively relaxed regarding the current task.
Therefore, $\epsilon$ plays a much less important role in ROGO.
We present the performance of different $\epsilon$ on CIFAR-100 Split in Figure~\ref{fig:res-cifar}-(d).
As shown in Figure~\ref{fig:res-cifar}-(d), the performance of GPM drops significantly when $\epsilon \geq 0.97$, the optimal value reported in GPM, while ROGO consistently performs well even with $\epsilon = 0.98$.
Generally, ROGO is more robust on the threshold $\epsilon$.

In brief, our approach universally outperforms selected baselines without extra data buffers.
Achieving better average accuracy, ROGO also improves forward knowledge transfer by exploring a larger optimization space than GPM.
To validate the efficiency of our relaxing strategy, we further compare ROGO-Exp with well-established and competitive expansion-based methods in the next section.

\subsection{Comparison with Expansion-based Methods}
\label{sec:4-3}

The above experiments exhibit the superiority of our approach when maintaining a fixed network capacity.
However, under the scenario with no limit on the network capacity, expansion-based methods achieve great performance by allocating new neurons or modules.
Therefore, to further validate our strategy, we compare ROGO and ROGO-Exp with relative expansion-based methods.

\begin{figure}[t]
    \centering  
    \includegraphics[width=\linewidth]{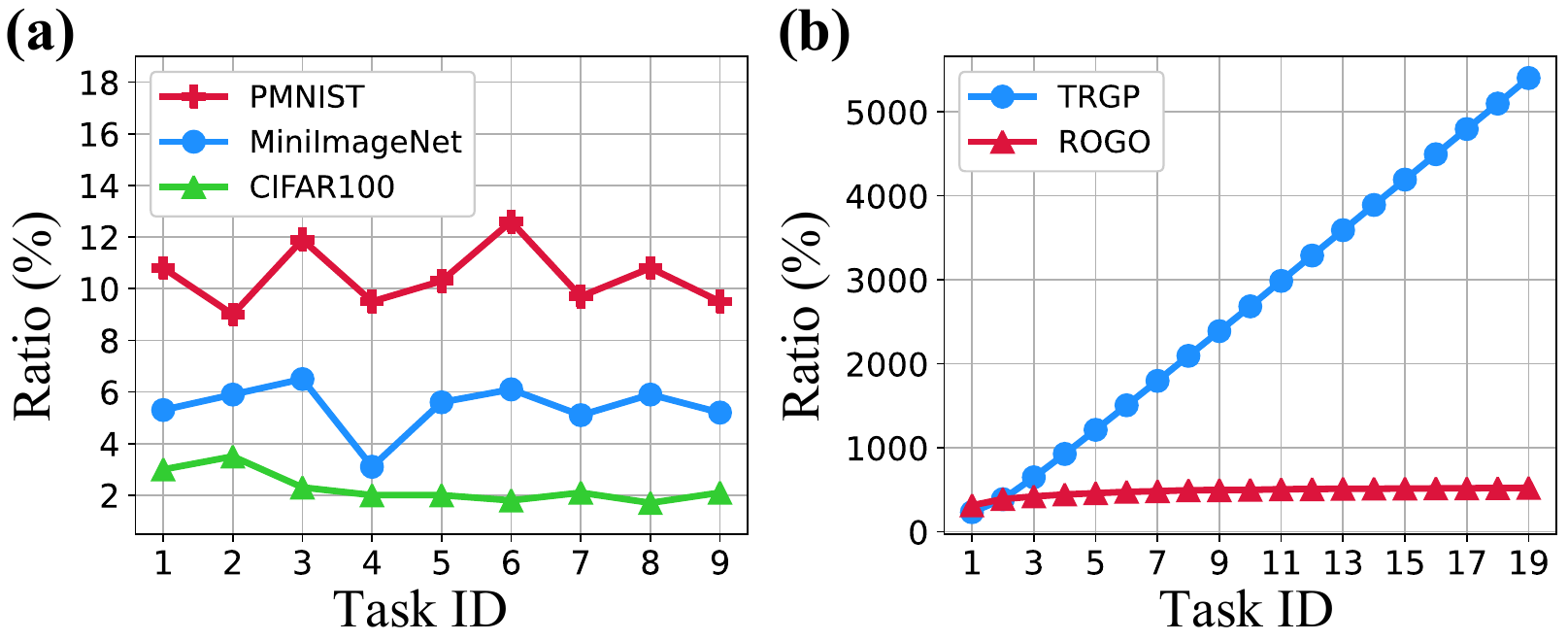}
    \caption{(a) Relaxing ratios of the last layer on CIFAR-100 Split, MiniImageNet, and PMNIST. (b) Ratios of the amount of extra parameters concerning the amount of the parameters of the network architecture on MiniImageNet.}  
    \label{fig:params}
    \vspace{-15pt}
\end{figure}

In this section, we adopt TRGP~\cite{lin2022trgp}, which expands the optimization space by retraining parameters within the selected trust regions, achieving superior performance.
In the inference phase, TRGP reuses the parameters in corresponding trust regions memorized after learning this task.
In contrast, GPM and our ROGO only store the representation of the frozen space.
Therefore, although indeed a stable network capacity is allocated for each task, the entire memory size of TRGP grows continually.
As shown in Figure~\ref{fig:params}-(b), after learning the last task on MiniImageNet, TRGP requires around 5,000\% extra parameters with respect to the network capacity.
Results on other benchmarks provided in Appendix~\ref{app:memory-usage} further substantiate that TRGP introduces a significant number of extra parameters.
Thus, we categorize TRGP as an expansion-based method here.

In this setting, the main difference between ROGO-Exp and TRGP is the strategy of deciding which part of the frozen space to reuse.
We conduct experiments on all four benchmarks and report the results in Table~\ref{tab:trgp-compare-exp}.
The percentages indicate the ratios of the rank of the relaxing space with respect to the corresponding frozen space.
We evaluate two constant ratios and further use the ratios in TRGP, denoted as T\%.
As TRGP selects the top 2 tasks as the trust regions, T\% is larger than 80\% at most times.
According to Table~\ref{tab:trgp-compare-exp}, IRGP-Exp universally outperforms TRGP, even by relaxing only 50\% of the frozen space.
Moreover, we observe that ROGO gains superior performance over both TRGP and ROGO-Exp on MiniImageNet.
We assume that the reason is the positive backward knowledge transfer.
Training the network on new tasks also benefits the previous task.
In General, our approach achieves better performance with a comparable size of the relaxing space, which substantiates the efficiency of our searching strategy.

We further modify TRGP as TRGP-Reg with similar regularization terms on the scale matrices as our ROGO to compare the relaxing strategies.
We report the results on four benchmarks with three representative regularization weights $\beta$ on TRGP-Reg in Table~\ref{tab:trgp-compare-non-exp}.
As shown in Table~\ref{tab:trgp-compare-non-exp}, ROGO significantly outperforms TRGP-Reg, especially on PMNIST, gaining around 20\% ACC improvement.
Generally, ROGO achieves better or comparable $\Omega_{new}$ than TRGP under or without the constraint of a fixed network capacity.
Detailed results are included in Appendix~\ref{app:others}.

\section{Analysis and Discussion}
\label{sec:5}

To gain a deeper insight into ROGO, we investigate the trend of scales of the space relaxed by our strategy.
With the theoretical upper bound of the rank of the relaxed subspace provided in Theorem~\ref{the:rank-k}, we further inspect the ratios of the relaxing spaces concerning corresponding frozen spaces in practice.
Results of the last layer on three different settings are provided in Figure~\ref{fig:params}-(a).
As shown in Figure~\ref{fig:params}-(a), relaxing ratios generally maintain a stable trend, fluctuating smoothly within a small range over sequential tasks on both benchmarks.
As different tasks explore different optimization directions, ideal relaxing spaces vary across tasks, in accordance with the fluctuation of our results.
To further investigate our framework, we provide the ablation study of related hyper-parameters on CIFAR-100 Split in Table~\ref{tab:hyper-param}.

\textbf{Ablation study of the relaxing ratios.}
ROGO mediates the stability-plasticity dilemma by controlling the dimension and flexibility of the relaxing space by $\gamma$ in Equation~(\ref{eq:criteria}).
For better illustration, here we represent $\gamma$ by $\zeta=\cos{\gamma}$.
We first conduct an ablation study of $\zeta$, reporting both the performance and the relaxing ratios in Table~\ref{tab:ablation-zeta}, where $\zeta_{conv}$ denotes the hyper-parameter for convolutional layers and $\zeta_{fc}$ denotes the hyper-parameter for fully connected layers.
As shown in Table~\ref{tab:ablation-zeta}, increasing $\zeta$ gradually constrains the scale of the relaxing space, thus alleviating the forgetting.
On the other hand, a small $\zeta$ guarantees sufficient forward knowledge transfer, while fails on consolidating previous knowledge.
Generally, our method persists in a superb performance with $\zeta \geq 0.80$.

\textbf{Ablation study of the regularization weights.}
Moreover, we observe the performance of ROGO with various regularization weights $\beta$ in Table~\ref{tab:ablation-beta}.
In this work, we adopt a consistent $\beta$ for the entire network.
According to Table~\ref{tab:ablation-beta}, 
Similarly, we observe less forgetting on larger $\beta$, which constrains the update of parameters within the relaxing space more strictly.
However, strict constraints also lead to limited performance on new tasks as discussed in Section~\ref{sec:4}.
In a nutshell, $\zeta$ and $\beta$ operate together in ROGO to overcome catastrophic forgetting and enhance forward transfer.

\textbf{Training time.}
The dimension of the frozen spaces keeps growing as the tasks accumulate, leading to expanding range for searching relaxing spaces.
Therefore, the computation complexity and time consumption are supposed to increase gradually.
We further reported the time consumption comparison on CIFAR-100 Split and MiniImageNet in Appendix~\ref{app:time-consume}.
According to Table~\ref{tab:time-comparison}, ROGO takes around 50\% more time than GPM, similar to TRGP.
In general, the practical efficiency of our approach is acceptable.

\section{Conclusion}

In this paper, we proposed ROGO, a novel continual learning framework that facilitates forward knowledge transfer in gradient projection methods within a fixed network capacity.
ROGO imposes a restricted orthogonal constraint allowing parameters to be updated in the direction oblique to the whole frozen space, which thus explores a larger optimization space.
Extensive experiments demonstrate that our ROGO framework surpasses related state-of-the-art approaches on diverse benchmarks. 
Moreover, we proposed ROGO-Exp that allows network expansion with the relaxing space, which achieves better performance than related expansion-based methods.
We also provided theoretical analysis validating the efficiency of our algorithm.




\nocite{langley00}

\bibliography{example_paper}

\begin{thebibliography}{48}
\providecommand{\natexlab}[1]{#1}
\providecommand{\url}[1]{\texttt{#1}}
\expandafter\ifx\csname urlstyle\endcsname\relax
  \providecommand{\doi}[1]{doi: #1}\else
  \providecommand{\doi}{doi: \begingroup \urlstyle{rm}\Url}\fi

\bibitem[Aljundi et~al.(2018)Aljundi, Babiloni, Elhoseiny, Rohrbach, and
  Tuytelaars]{aljundi2018memory}
Aljundi, R., Babiloni, F., Elhoseiny, M., Rohrbach, M., and Tuytelaars, T.
\newblock Memory aware synapses: Learning what (not) to forget.
\newblock In \emph{Proceedings of the European Conference on Computer Vision
  (ECCV)}, pp.\  139--154, 2018.

\bibitem[Bennani et~al.(2020)Bennani, Doan, and
  Sugiyama]{bennani2020generalisation}
Bennani, M.~A., Doan, T., and Sugiyama, M.
\newblock Generalisation guarantees for continual learning with orthogonal
  gradient descent.
\newblock \emph{arXiv preprint arXiv:2006.11942}, 2020.

\bibitem[Bulatov(2011)]{NotMNIST}
Bulatov, Y.
\newblock Notmnist dataset.
\newblock \url{http://yaroslavvb.com/upload/notMNIST/}, 2011.

\bibitem[Chaudhry et~al.(2018)Chaudhry, Ranzato, Rohrbach, and
  Elhoseiny]{chaudhry2018efficient}
Chaudhry, A., Ranzato, M., Rohrbach, M., and Elhoseiny, M.
\newblock Efficient lifelong learning with a-gem.
\newblock \emph{arXiv preprint arXiv:1812.00420}, 2018.

\bibitem[Chaudhry et~al.(2019)Chaudhry, Rohrbach, Elhoseiny, Ajanthan, Dokania,
  Torr, and Ranzato]{chaudhry2019continual}
Chaudhry, A., Rohrbach, M., Elhoseiny, M., Ajanthan, T., Dokania, P.~K., Torr,
  P.~H., and Ranzato, M.
\newblock Continual learning with tiny episodic memories.
\newblock \emph{CoRR}, abs/1902.10486, 2019.

\bibitem[Chenshen et~al.(2018)Chenshen, HERRANZ, Xialei,
  et~al.]{chenshen2018memory}
Chenshen, W., HERRANZ, L., Xialei, L., et~al.
\newblock Memory replay gans: Learning to generate images from new categories
  without forgetting [c].
\newblock In \emph{The 32nd International Conference on Neural Information
  Processing Systems, Montr{\'e}al, Canada}, pp.\  5966--5976, 2018.

\bibitem[Choi et~al.(2021)Choi, El-Khamy, and Lee]{choi2021dual}
Choi, Y., El-Khamy, M., and Lee, J.
\newblock Dual-teacher class-incremental learning with data-free generative
  replay.
\newblock In \emph{Proceedings of the IEEE/CVF Conference on Computer Vision
  and Pattern Recognition}, pp.\  3543--3552, 2021.

\bibitem[Cong et~al.(2020)Cong, Zhao, Li, Wang, and Carin]{cong2020gan}
Cong, Y., Zhao, M., Li, J., Wang, S., and Carin, L.
\newblock Gan memory with no forgetting.
\newblock \emph{Advances in Neural Information Processing Systems},
  33:\penalty0 16481--16494, 2020.

\bibitem[De~Lange et~al.(2021)De~Lange, Aljundi, Masana, Parisot, Jia,
  Leonardis, Slabaugh, and Tuytelaars]{de2021continual}
De~Lange, M., Aljundi, R., Masana, M., Parisot, S., Jia, X., Leonardis, A.,
  Slabaugh, G., and Tuytelaars, T.
\newblock A continual learning survey: Defying forgetting in classification
  tasks.
\newblock \emph{IEEE transactions on pattern analysis and machine
  intelligence}, 44\penalty0 (7):\penalty0 3366--3385, 2021.

\bibitem[Douillard et~al.(2022)Douillard, Ram{\'e}, Couairon, and
  Cord]{douillard2022dytox}
Douillard, A., Ram{\'e}, A., Couairon, G., and Cord, M.
\newblock Dytox: Transformers for continual learning with dynamic token
  expansion.
\newblock In \emph{Proceedings of the IEEE/CVF Conference on Computer Vision
  and Pattern Recognition}, pp.\  9285--9295, 2022.

\bibitem[Ebrahimi et~al.(2020)Ebrahimi, Meier, Calandra, Darrell, and
  Rohrbach]{ebrahimi2020adversarial}
Ebrahimi, S., Meier, F., Calandra, R., Darrell, T., and Rohrbach, M.
\newblock Adversarial continual learning.
\newblock In \emph{European Conference on Computer Vision}, pp.\  386--402.
  Springer, 2020.

\bibitem[Ehret et~al.(2020)Ehret, Henning, Cervera, Meulemans, Von~Oswald, and
  Grewe]{ehret2020continual}
Ehret, B., Henning, C., Cervera, M.~R., Meulemans, A., Von~Oswald, J., and
  Grewe, B.~F.
\newblock Continual learning in recurrent neural networks.
\newblock \emph{arXiv preprint arXiv:2006.12109}, 2020.

\bibitem[Farajtabar et~al.(2020)Farajtabar, Azizan, Mott, and
  Li]{farajtabar2020orthogonal}
Farajtabar, M., Azizan, N., Mott, A., and Li, A.
\newblock Orthogonal gradient descent for continual learning.
\newblock In \emph{International Conference on Artificial Intelligence and
  Statistics}, pp.\  3762--3773. PMLR, 2020.

\bibitem[Finn et~al.(2019)Finn, Rajeswaran, Kakade, and Levine]{finn2019online}
Finn, C., Rajeswaran, A., Kakade, S., and Levine, S.
\newblock Online meta-learning.
\newblock In \emph{International Conference on Machine Learning}, pp.\
  1920--1930. PMLR, 2019.

\bibitem[French(1997)]{french1997pseudo}
French, R.~M.
\newblock Pseudo-recurrent connectionist networks: An approach to
  the'sensitivity-stability'dilemma.
\newblock \emph{Connection Science}, 9\penalty0 (4):\penalty0 353--380, 1997.

\bibitem[Griffiths \& Ghahramani(2011)Griffiths and
  Ghahramani]{griffiths2011indian}
Griffiths, T.~L. and Ghahramani, Z.
\newblock The indian buffet process: An introduction and review.
\newblock \emph{Journal of Machine Learning Research}, 12\penalty0 (4), 2011.

\bibitem[Kao et~al.(2021)Kao, Jensen, van~de Ven, Bernacchia, and
  Hennequin]{kao2021natural}
Kao, T.-C., Jensen, K., van~de Ven, G., Bernacchia, A., and Hennequin, G.
\newblock Natural continual learning: success is a journey, not (just) a
  destination.
\newblock \emph{Advances in Neural Information Processing Systems},
  34:\penalty0 28067--28079, 2021.

\bibitem[Kemker et~al.(2018)Kemker, McClure, Abitino, Hayes, and
  Kanan]{kemker2018measuring}
Kemker, R., McClure, M., Abitino, A., Hayes, T., and Kanan, C.
\newblock Measuring catastrophic forgetting in neural networks.
\newblock In \emph{Proceedings of the AAAI Conference on Artificial
  Intelligence}, volume~32, 2018.

\bibitem[Kirkpatrick et~al.(2017)Kirkpatrick, Pascanu, Rabinowitz, Veness,
  Desjardins, Rusu, Milan, Quan, Ramalho, Grabska-Barwinska,
  et~al.]{kirkpatrick2017overcoming}
Kirkpatrick, J., Pascanu, R., Rabinowitz, N., Veness, J., Desjardins, G., Rusu,
  A.~A., Milan, K., Quan, J., Ramalho, T., Grabska-Barwinska, A., et~al.
\newblock Overcoming catastrophic forgetting in neural networks.
\newblock \emph{Proceedings of the national academy of sciences}, 114\penalty0
  (13):\penalty0 3521--3526, 2017.

\bibitem[Kong et~al.(2022)Kong, Liu, Wang, and Tao]{kong2022balancing}
Kong, Y., Liu, L., Wang, Z., and Tao, D.
\newblock Balancing stability and plasticity through advanced null space in
  continual learning.
\newblock \emph{arXiv preprint arXiv:2207.12061}, 2022.

\bibitem[Krizhevsky \& Hinton(2009)Krizhevsky and
  Hinton]{krizhevsky2009learning}
Krizhevsky, A. and Hinton, G.
\newblock Learning multiple layers of features from tiny images.
\newblock Technical report, Citeseer, 2009.

\bibitem[Kumar et~al.(2021)Kumar, Chatterjee, and Rai]{kumar2021bayesian}
Kumar, A., Chatterjee, S., and Rai, P.
\newblock Bayesian structural adaptation for continual learning.
\newblock In \emph{International Conference on Machine Learning}, pp.\
  5850--5860. PMLR, 2021.

\bibitem[Kurle et~al.(2019)Kurle, Cseke, Klushyn, Van Der~Smagt, and
  G{\"u}nnemann]{kurle2019continual}
Kurle, R., Cseke, B., Klushyn, A., Van Der~Smagt, P., and G{\"u}nnemann, S.
\newblock Continual learning with bayesian neural networks for non-stationary
  data.
\newblock In \emph{International Conference on Learning Representations}, 2019.

\bibitem[LeCun et~al.(1998)LeCun, Bottou, Bengio, and
  Haffner]{lecun1998gradient}
LeCun, Y., Bottou, L., Bengio, Y., and Haffner, P.
\newblock Gradient-based learning applied to document recognition.
\newblock \emph{Proceedings of the IEEE}, 86\penalty0 (11):\penalty0
  2278--2324, 1998.

\bibitem[Lin et~al.(2022)Lin, Yang, Fan, and Zhang]{lin2022trgp}
Lin, S., Yang, L., Fan, D., and Zhang, J.
\newblock Trgp: Trust region gradient projection for continual learning.
\newblock \emph{arXiv preprint arXiv:2202.02931}, 2022.

\bibitem[Liu \& Liu(2022)Liu and Liu]{liu2022continual}
Liu, H. and Liu, H.
\newblock Continual learning with recursive gradient optimization.
\newblock \emph{arXiv preprint arXiv:2201.12522}, 2022.

\bibitem[Lopez-Paz \& Ranzato(2017)Lopez-Paz and Ranzato]{lopez2017gradient}
Lopez-Paz, D. and Ranzato, M.
\newblock Gradient episodic memory for continual learning.
\newblock \emph{Advances in neural information processing systems}, 30, 2017.

\bibitem[Mallya \& Lazebnik(2018)Mallya and Lazebnik]{mallya2018packnet}
Mallya, A. and Lazebnik, S.
\newblock Packnet: Adding multiple tasks to a single network by iterative
  pruning.
\newblock In \emph{Proceedings of the IEEE conference on Computer Vision and
  Pattern Recognition}, pp.\  7765--7773, 2018.

\bibitem[McCloskey \& Cohen(1989)McCloskey and
  Cohen]{mccloskey1989catastrophic}
McCloskey, M. and Cohen, N.~J.
\newblock Catastrophic interference in connectionist networks: The sequential
  learning problem.
\newblock In \emph{Psychology of learning and motivation}, volume~24, pp.\
  109--165. Elsevier, 1989.

\bibitem[Mirzadeh et~al.(2020)Mirzadeh, Farajtabar, Pascanu, and
  Ghasemzadeh]{mirzadeh2020understanding}
Mirzadeh, S.~I., Farajtabar, M., Pascanu, R., and Ghasemzadeh, H.
\newblock Understanding the role of training regimes in continual learning.
\newblock \emph{Advances in Neural Information Processing Systems},
  33:\penalty0 7308--7320, 2020.

\bibitem[Netzer et~al.(2011)Netzer, Wang, Coates, Bissacco, Wu, and
  Ng]{netzer2011reading}
Netzer, Y., Wang, T., Coates, A., Bissacco, A., Wu, B., and Ng, A.~Y.
\newblock Reading digits in natural images with unsupervised feature learning.
\newblock In \emph{NIPS Workshop on Deep Learning and Unsupervised Feature
  Learning}, 2011.

\bibitem[Ng \& Winkler(2014)Ng and Winkler]{ng2014data}
Ng, H.-W. and Winkler, S.
\newblock A data-driven approach to cleaning large face datasets.
\newblock In \emph{2014 IEEE international conference on image processing
  (ICIP)}, pp.\  343--347. IEEE, 2014.

\bibitem[Parisi et~al.(2019)Parisi, Kemker, Part, Kanan, and
  Wermter]{parisi2019continual}
Parisi, G.~I., Kemker, R., Part, J.~L., Kanan, C., and Wermter, S.
\newblock Continual lifelong learning with neural networks: A review.
\newblock \emph{Neural Networks}, 113:\penalty0 54--71, 2019.

\bibitem[Ramesh \& Chaudhari(2021)Ramesh and Chaudhari]{ramesh2021model}
Ramesh, R. and Chaudhari, P.
\newblock Model zoo: A growing brain that learns continually.
\newblock In \emph{International Conference on Learning Representations}, 2021.

\bibitem[Ratcliff(1990)]{ratcliff1990connectionist}
Ratcliff, R.
\newblock Connectionist models of recognition memory: constraints imposed by
  learning and forgetting functions.
\newblock \emph{Psychological review}, 97\penalty0 (2):\penalty0 285, 1990.

\bibitem[Rusu et~al.(2016)Rusu, Rabinowitz, Desjardins, Soyer, Kirkpatrick,
  Kavukcuoglu, Pascanu, and Hadsell]{rusu2016progressive}
Rusu, A.~A., Rabinowitz, N.~C., Desjardins, G., Soyer, H., Kirkpatrick, J.,
  Kavukcuoglu, K., Pascanu, R., and Hadsell, R.
\newblock Progressive neural networks.
\newblock \emph{arXiv preprint arXiv:1606.04671}, 2016.

\bibitem[Saha et~al.(2021)Saha, Garg, and Roy]{saha2021gradient}
Saha, G., Garg, I., and Roy, K.
\newblock Gradient projection memory for continual learning.
\newblock \emph{arXiv preprint arXiv:2103.09762}, 2021.

\bibitem[Serra et~al.(2018)Serra, Suris, Miron, and
  Karatzoglou]{serra2018overcoming}
Serra, J., Suris, D., Miron, M., and Karatzoglou, A.
\newblock Overcoming catastrophic forgetting with hard attention to the task.
\newblock In \emph{International Conference on Machine Learning}, pp.\
  4548--4557. PMLR, 2018.

\bibitem[Shin et~al.(2017)Shin, Lee, Kim, and Kim]{shin2017continual}
Shin, H., Lee, J.~K., Kim, J., and Kim, J.
\newblock Continual learning with deep generative replay.
\newblock \emph{Advances in neural information processing systems}, 30, 2017.

\bibitem[Stallkamp et~al.(2011)Stallkamp, Schlipsing, Salmen, and
  Igel]{stallkamp2011german}
Stallkamp, J., Schlipsing, M., Salmen, J., and Igel, C.
\newblock The german traffic sign recognition benchmark: a multi-class
  classification competition.
\newblock In \emph{The 2011 international joint conference on neural networks},
  pp.\  1453--1460. IEEE, 2011.

\bibitem[Teng et~al.(2022)Teng, Choromanska, Campbell, Lu, Ram, and
  Horesh]{tengovercoming}
Teng, Y., Choromanska, A., Campbell, M., Lu, S., Ram, P., and Horesh, L.
\newblock Overcoming catastrophic forgetting via direction-constrained
  optimization.
\newblock In \emph{European Conference on Machine Learning and Principles and
  Practice of Knowledge Discovery in Databases}, 2022.

\bibitem[Thrun \& Mitchell(1995)Thrun and Mitchell]{thrun1995lifelong}
Thrun, S. and Mitchell, T.~M.
\newblock Lifelong robot learning.
\newblock \emph{Robotics and autonomous systems}, 15\penalty0 (1-2):\penalty0
  25--46, 1995.

\bibitem[Vinyals et~al.(2016)Vinyals, Blundell, Lillicrap, Wierstra,
  et~al.]{vinyals2016matching}
Vinyals, O., Blundell, C., Lillicrap, T., Wierstra, D., et~al.
\newblock Matching networks for one shot learning.
\newblock \emph{Advances in neural information processing systems}, 29, 2016.

\bibitem[Wang et~al.(2021)Wang, Li, Sun, and Xu]{wang2021training}
Wang, S., Li, X., Sun, J., and Xu, Z.
\newblock Training networks in null space of feature covariance for continual
  learning.
\newblock In \emph{Proceedings of the IEEE/CVF Conference on Computer Vision
  and Pattern Recognition}, pp.\  184--193, 2021.

\bibitem[Xiao et~al.(2017)Xiao, Rasul, and Vollgraf]{xiao2017fashion}
Xiao, H., Rasul, K., and Vollgraf, R.
\newblock Fashion-mnist: a novel image dataset for benchmarking machine
  learning algorithms.
\newblock \emph{arXiv preprint arXiv:1708.07747}, 2017.

\bibitem[Yoon et~al.(2017)Yoon, Yang, Lee, and Hwang]{yoon2017lifelong}
Yoon, J., Yang, E., Lee, J., and Hwang, S.~J.
\newblock Lifelong learning with dynamically expandable networks.
\newblock \emph{arXiv preprint arXiv:1708.01547}, 2017.

\bibitem[Yoon et~al.(2019)Yoon, Kim, Yang, and Hwang]{yoon2019scalable}
Yoon, J., Kim, S., Yang, E., and Hwang, S.~J.
\newblock Scalable and order-robust continual learning with additive parameter
  decomposition.
\newblock \emph{arXiv preprint arXiv:1902.09432}, 2019.

\bibitem[Zeng et~al.(2019)Zeng, Chen, Cui, and Yu]{zeng2019continual}
Zeng, G., Chen, Y., Cui, B., and Yu, S.
\newblock Continual learning of context-dependent processing in neural
  networks.
\newblock \emph{Nature Machine Intelligence}, 1\penalty0 (8):\penalty0
  364--372, 2019.

\end{thebibliography}
\bibliographystyle{icml2023}

\newpage
\appendix
\onecolumn

\section{Proof}

In this section, we sequentially provide the proof of Theorem~\ref{the:max-rank} and~\ref{the:rank-k}.
For better comprehension, we first introduce complementary Lemma~\ref{lem:angle}.
For simplification, all vectors here are assumed to be unit vectors, namely $\Vert v \Vert = 1$.

\subsection{Proof of Lemma~\ref{lem:angle}}
\label{app:proof-lem-angle}

\begin{lemma}
Denote the relaxed space as $V = \mathrm{span}\{v_1, v_2, ..., v_n\}$, where $v_n$ is the last base included in $V$.
Given representation space $U$, $\forall v \in V$, we have $\mathit{\Theta} ( v, U ) \leq \mathit{\Theta} ( v_n, U )$.
\label{lem:angle}
\end{lemma}

Depicting the angle by projection form, Lemma~\ref{lem:angle} can be expressed as:

\begin{equation}
    \forall v \in V, \Vert \mathrm{Proj}_{U} (v) \Vert \geq \Vert \mathrm{Proj}_{U} (v_n) \Vert
\end{equation}

Denote $\mathbf{B} = [u_1, ..., u_m]$ as the representative matrix of the representation space $U = \mathrm{span}\{u_1, ..., u_m\}$, where $u_i$ is the $i$-th normalized orthogonal base of $U$.
Lemma~\ref{lem:angle} can be further expressed as:

\begin{equation}
    \forall v \in V, v^T \mathbf{B} \mathbf{B}^T v \geq v^T_n \mathbf{B} \mathbf{B}^T v_n
\end{equation}

As $v_i$s are the basis sequentially appended by our searching strategy, for any $i \leq j$, we have:

\begin{equation}
    v^T_i \mathbf{B} \mathbf{B}^T v_i \geq v^T_j \mathbf{B} \mathbf{B}^T v_j
    \label{eq:base-proj}
\end{equation}

Consider a special case where the current relaxed subspace has only one base, denoted by $V = \mathrm{span}\{v_1\}$, namely $n=1$.
Obviously, Lemma~\ref{lem:angle} is true in this case.
Thus, we consider the general case that $n \geq 2$.

We provide proof by inductive reasoning. 
Note that as we assume all vectors to be unit vectors, we have $\sum_i \Vert w_i \Vert^2_2 = 1$.

First, we consider a special case $V = \mathrm{span}\{v_1, v_2\}$, namely $n=2$.
Assume there exists $v = w_1 v_1 + w_2 v_2$ that $v^T \mathbf{B} \mathbf{B}^T v < v^T_2 \mathbf{B} \mathbf{B}^T v_2$, with $w_1^2 + w_2^2 = 1$, we have:

\begin{equation}
    w_1^2 v^T_2 \mathbf{B} \mathbf{B}^T v_2 > w_1^2 v^T_1 \mathbf{B} \mathbf{B}^T v_1 + 2 w_1 w_2 v^T_1 \mathbf{B} \mathbf{B}^T v_2
\end{equation}

Construct $v' = w_2 v_1 - w_1 v_2$, we have:

\begin{equation}
    \begin{aligned}
    v'^T \mathbf{B} \mathbf{B}^T v'&= w_2^2 v^T_1 \mathbf{B} \mathbf{B}^T v_1 + w_1^2 v^T_2 \mathbf{B} \mathbf{B}^T v_2 - 2 w_1 w_2 v^T_1 \mathbf{B} \mathbf{B}^T v_2 \\
    &> w_2^2 v^T_1 \mathbf{B} \mathbf{B}^T v_1 + w_1^2 v^T_2 \mathbf{B} \mathbf{B}^T v_2 + w_1^2 v^T_1 \mathbf{B} \mathbf{B}^T v_1 - w_1^2 v^T_2 \mathbf{B} \mathbf{B}^T v_2 \\
    &= v^T_1 \mathbf{B} \mathbf{B}^T v_1
    \end{aligned}
\end{equation}

which contradicts $\mathit{\Theta} ( v_1, U ) \leq \mathit{\Theta} ( v, U )$ for $\forall v \in \mathrm{span}\{v_1, v_2\}$.

Then we consider the general case $V = \mathrm{span}\{v_1, ..., v_t\}$.
For $\forall v \in V$, we have $\mathit{\Theta} ( v, U ) \leq \mathit{\Theta} ( v_t, U )$.
After $v_{t+1}$ is included, we assume that there exists $v \in V$ that $\mathit{\Theta} ( v, U ) > \mathit{\Theta} ( v_{t+1}, U )$.
Then we can find the maximum $s$ satisfying that there exists $v = \sum_{i=1}^s w_i v_i + w_{t+1} v_{t+1}$ that $\mathit{\Theta} ( v, U ) > \mathit{\Theta} ( v_{t+1}, U )$ and for $\forall v' = \sum_{i=1}^{s-1} w_i v_i + w_{t+1} v_{t+1}$, we have $\mathit{\Theta} ( v', U ) \leq \mathit{\Theta} ( v_{t+1}, U )$.
When $s = 1$, it is similar to the special case, so the proof is omitted.
Thus, we consider the case where $s \geq 2$.
For simplification, we express $v$ as $v = c_0 v_0 + c_1 v_s + c_2 v_{t+1}$ with $v_0 = \sum_{i=1}^{s-1} a_i v_i$, where $c_i$s and $a_i$s are coefficients.
We have:

\begin{equation}
    \begin{aligned}
    (c_0 v_0 + c_1 v_s + c_2 v_{t+1})^T \mathbf{B} \mathbf{B}^T (c_0 v_0 + c_1 v_s + c_2 v_{t+1}) < v^T_{t+1} \mathbf{B} \mathbf{B}^T v_{t+1}
    \end{aligned}
    \label{eq:proj-general}
\end{equation}

As $\mathit{\Theta} ( w_1 v_0 + w_2 v_s, U ) \leq \mathit{\Theta} ( v_s, U ) \leq \mathit{\Theta} ( v_{t+1}, U )$, we have:

\begin{equation}
    (w_1 v_0 + w_2 v_s)^T \mathbf{B} \mathbf{B}^T (w_1 v_0 + w_2 v_s)\geq v_s^T \mathbf{B} \mathbf{B}^T v_s
\end{equation}

which is:

\begin{equation}
    w_1^2 v_0^T \mathbf{B} \mathbf{B}^T v_0 + 2 w_1 w_2 v_0^T \mathbf{B} \mathbf{B}^T v_s \geq w_1^2 v_s^T \mathbf{B} \mathbf{B}^T v_s \geq w_1^2 v_t^T \mathbf{B} \mathbf{B}^T v_t
\end{equation}

Similarly we have:

\begin{equation}
    w_1^2 v_0^T \mathbf{B} \mathbf{B}^T v_0 + 2 w_1 w_2 v_0^T \mathbf{B} \mathbf{B}^T v_t \geq w_1^2 v_t^T \mathbf{B} \mathbf{B}^T v_t
\end{equation}

Then we can express Equation~(\ref{eq:proj-general}) as:

\begin{equation}
    v^T_{t+1} \mathbf{B} \mathbf{B}^T v_{t+1} > (c_0^2 + c_2^2) v_t^T \mathbf{B} \mathbf{B}^T v_t + c_1^2 v_s^T \mathbf{B} \mathbf{B}^T v_s + 2 c_1 c_2 v_s^T \mathbf{B} \mathbf{B}^T v_t
\end{equation}

As $\Vert v \Vert = \Vert v_i \Vert = 1$, $c_0^2 + c_1^2 + c_2^2 = 1$.
Then we have:

\begin{equation}
    - 2 c_1 c_2 v_s^T \mathbf{B} \mathbf{B}^T v_t > c_1^2 v_s^T \mathbf{B} \mathbf{B}^T v_s - c_1^2 v_{t+1}^T \mathbf{B} \mathbf{B}^T v_{t+1}
\end{equation}

Construct $v' = \frac{c_2 v_s - c_1 v_{t+1}}{\sqrt{c_1^2 + c_2^2}}$, we have:

\begin{equation}
    \begin{aligned}
    v'^T \mathbf{B} \mathbf{B}^T v'&= \frac{1}{c_1^2 + c_2^2} (c_2^2 v_s^T \mathbf{B} \mathbf{B}^T v_s + c_1^2 v_{t+1}^T \mathbf{B} \mathbf{B}^T v_{t+1} - 2 c_1 c_2 v_s^T \mathbf{B} \mathbf{B}^T v_t)\\
    &> \frac{1}{c_1^2 + c_2^2} (c_2^2 v_s^T \mathbf{B} \mathbf{B}^T v_s + c_1^2 v_{t+1}^T \mathbf{B} \mathbf{B}^T v_{t+1} + c_1^2 v_s^T \mathbf{B} \mathbf{B}^T v_s - c_1^2 v_{t+1}^T \mathbf{B} \mathbf{B}^T v_{t+1})\\
    &= v_s^T \mathbf{B} \mathbf{B}^T v_s
    \end{aligned}
\end{equation}

which contradicts $\mathit{\Theta} ( v_s, U ) \leq \mathit{\Theta} ( v, U )$ for $\forall v \in \mathrm{span}\{v_s, v_{s+1}, ..., v_t, v_{t+1}\}$.

Thus, for $\forall v \in V = \mathrm{span}\{v_1, ..., v_n\}$, we have $\mathit{\Theta} ( v, U ) \leq \mathit{\Theta} ( v_n, U )$.

\subsection{Proof of Theorem~\ref{the:max-rank}}
\label{app:proof-max-rank}

Here we provide proof of Theorem~\ref{the:rank-k}.
Denote the whole solution set as $S=\{u \vert \mathit{\Theta} ( u, U ) \leq \gamma \ \mathrm{and} \ u \in U^f\}$ where $U$ is the representation space, $U^f$ is the frozen space and $\gamma$ is the threshold.
Theorem~\ref{the:max-rank} can be expressed as that all subspace $V' \subseteq S$ satisfies $\mathrm{dim} (V') \leq \mathrm{dim} (V)$, where $V$ is the relaxing space obtained by our searching strategy.

For the sake of contradiction, we assume there exists $V' \subseteq S$ that $\mathrm{dim} (V') > \mathrm{dim} (V)$.
Denote $V^c_f$ as the orthogonal complement of $V$ with respect to the frozen space $U^f$, as the relaxing space is a subspace of the frozen space.
According to Lemmoa~\ref{lem:angle}, for $\forall v \in V^c_f$, we have $\mathit{\Theta} ( v, U ) > \gamma$.
We also have:

\begin{equation}
\mathrm{dim} (V' \cap V^c_f)= \mathrm{dim} (V') + \mathrm{dim} (V^c_f) - \mathrm{dim} (V' + V^c_f) > 0
\end{equation}

Thus, there exists $v' \in V'$ that $v' \in U^c_f$, namely there exists $v' \in S$ that $\mathit{\Theta} ( v', U ) > \gamma$, which is contradict.
Therefore, the relaxing space obtained by our searching strategy takes up the maximum subspace of the whole solution set.

\subsection{Proof of Theorem~\ref{the:rank-k}}
\label{app:proof-rank-k}

Here we provide proof of Theorem~\ref{the:rank-k}.
Denote the relaxing space and the representation space as $V$ and $U = \mathrm{span}\{u_1, ..., u_{k_r}\}$ respectively.
With Lemma~\ref{lem:angle}, we have $\forall v \in V$, $\mathit{\Theta} ( v, U ) \leq \mathit{\Theta} ( v_{k_v}, U ) < \frac{\pi}{2}$, where $k_v$ denotes $\mathrm{dim} (V)$, the dimension of $V$.
In other words, 

\begin{equation}
    \forall v \in V, \ v \not \perp U
    \label{eq:no-vert}
\end{equation}

Similar to Theorem~\ref{the:rank-k}, assume the dimension of $V$ is larger than $k_r$, namely $\mathrm{dim} (V) > \mathrm{dim} (U) = k_r$.
Denote $U^c$ as the orthogonal complement of $U$ with respect to the whole space $\mathbb{R}$.
Obviously, $\mathrm{dim} (U^c) = n-k_r$, where $n$ is the dimension of $\mathbb{R}$.
Then we have:

\begin{equation}
    \begin{aligned}
    \mathrm{dim} (V \cap U^c)&= \mathrm{dim} (V) + \mathrm{dim} (U^c) - \mathrm{dim} (V + U^c)\\
    &> k_r + (n-k_r) - n = 0
    \end{aligned}
\end{equation}

Thus, there exists $v' \in V$ such that $v' \in U^c$ too.
As $U^c$ is the orthogonal complement, $\forall u \in U^c$, $u \perp U$.
Then we have $v' \perp U$, which contradicts Equation~(\ref{eq:no-vert}).
Therefore, the assumption is aborted.
We have that $k_v \leq k_r$.
The upper bound of the dimension of $V$ is $k_r$, namely the dimension of the representation space $U$.

\section{Experimental Setup}
\label{app:exp-setup}

\subsection{Datasets}
\label{app:dataset-stat}

Here we introduce the datasets we use for evaluation.
\textbf{1) CIFAR-100 Split}
\citet{saha2021gradient} constructed \textbf{CIFAR-100 Split}, by splitting CIFAR100~\citep{krizhevsky2009learning} into 10 tasks where each task has 10 classes.
\textbf{2) MiniImageNet}
Following \citet{saha2021gradient}, we split MiniImageNet~\citep{vinyals2016matching} into 20 sequential tasks with 5 classes each.
\textbf{3) Permuted MNIST (PMNIST)}
PMNIST~\citep{kirkpatrick2017overcoming} is a variant of MNIST~\citep{lecun1998gradient} where each task has a different permutation of inputting images, consists of 10 sequential tasks with 10 classes each. 
\textbf{4) Mixture}
\citet{serra2018overcoming} first proposed Mixture consisting of 8 datasets, including CIFAR-10~\citep{krizhevsky2009learning}, MNIST~\citep{lecun1998gradient}, CIFAR-100~\citep{krizhevsky2009learning}, SVHN~\citep{netzer2011reading}, FashionMNIST~\citep{xiao2017fashion}, TrafficSigns~\citep{stallkamp2011german}, FaceScrub~\citep{ng2014data}, and  NotMNIST~\citep{NotMNIST}, from which \citet{ebrahimi2020adversarial} further constructed 5-Datasets.
Here we follow the original harder benchmark.
Particularly, we consider all tasks as a sequence except TrafficSigns~\citep{stallkamp2011german}, which we failed to access.
\textbf{5) CIFAR-100 Sup} 
In addition, following \citet{yoon2019scalable}, we adopt \textbf{CIFAR-100 Sup} consisting of 20 superclasses as sequential tasks.
Here we adopt the five different predefined orders proposed by \citet{yoon2019scalable} and report the main results in Appendix~\ref{app:others}.
Among all evaluated datasets, PMNIST is a benchmark under the domain-incremental scenario, while other four datasets are under the task-incremental scenario.

Moreover, we provide the statistics of selected datasets in Table~\ref{tab:data-stat-3} and Table~\ref{tab:stat-mixture}.
For the Mixture benchmark, the images of MNIST, FashionMNIST, and notMNIST are replicated across all RGB channels following \citet{serra2018overcoming}.

\begin{table}[h]
    \centering
    \caption{Statistics of CIFAR-100 Split, MiniImageNet, and PMNIST.}
    \begin{tabular}{lcccc}
        \toprule
         & CIFAR-100 Split & CIFAR-100 Sup & MiniImageNet & PMNIST \\
        \midrule
        Image Size & $32 \times 32$ & $32 \times 32$ & $84 \times 84$ & $28 \times 28$ \\
        Channels & 3 & 3 & 3 & 1 \\
        Classes & 100 & 100 & 100 & 10 \\
        Tasks & 10 & 20 & 20 & 10 \\
        Classes/task & 10 & 5 & 5 & 10 \\
        Training Samples/task & 4,750 & 2,375 & 2,375 & 54,000 \\
        Validation Samples/task & 250 & 125 & 125 & 6,000 \\
        Testing Samples/task & 1,000 & 500 & 500 & 10,000 \\
        \bottomrule
    \end{tabular}
    \label{tab:data-stat-3}
\end{table}

\begin{table}[h]
    \centering
    \caption{Statistics of Mixture benchmark.}
    \begin{tabular}{lrrrr}
        \toprule
        Dataset & Classes & \# Taining & \# Validation & \# Testing \\
        \midrule
        CIFAR-10~\citep{krizhevsky2009learning} & 10 & 47,500 & 2,500 & 10,000 \\
        MNIST~\citep{lecun1998gradient} & 10 & 57,000 & 3,000 & 10,000 \\
        CIFAR-100~\citep{krizhevsky2009learning} & 100 & 47,500 & 2,500 & 10,000 \\
        SVHN~\citep{netzer2011reading} & 10 & 69,595 & 3,662 & 26,032 \\ 
        FashionMNIST~\citep{xiao2017fashion} & 10 & 57,000 & 3,000 & 10,000 \\
        FaceScrub~\citep{ng2014data} & 100 & 19,570 & 1,030 & 2,289 \\
        NotMNIST~\citep{NotMNIST} & 10 & 16,011 & 842 & 1,873 \\
        \bottomrule
    \end{tabular}
    \label{tab:stat-mixture}
\end{table}

\subsection{Model Details}
\label{app:model-detail}

\textbf{MLP architecture:}
We adopt a 3-layer model including two hidden layers with 100 neurons each for the PMNIST setting, the same as \citet{lopez2017gradient}.
ReLU is used as the activate function here and for all other architectures.
Also, we use softmax with cross entropy loss on all settings.

\textbf{AlexNet architecture:}
For CIFAR-100 Split setting, we adopt the same network as \citet{serra2018overcoming} with batch normalization, including two fully connected layers and three convolutional layers.
The convolutional layers have $4 \times 4$, $3 \times 3$, and $2 \times 2$ kernel sizes with 64, 128, and 256 filters respectively.
After each convolutional layer, we add batch normalization and $2 \times 2$ max-pooling.
Each fully connected layer has 2048 units.
For the first two layers, we use the dropout of 0.2, and for the rest layers, we use the dropout of 0.5.

\textbf{Modified LeNet-5 architecture:}
For the CIFAR-100 Sup setting, a modified LeNet-5 architecture consisting of two convolutional layers and two fully connected layers is adopted, similar to \citet{saha2021gradient}.
Max-pooling of $3 \times 2$ is used after each convolutional layer.
The last two layers have 800 and 500 units respectively.

\textbf{Reduced ResNet-18 architecture:}
We adopt the same reduced ResNet-18 architecture as \citet{saha2021gradient} for the MiniImageNet and Mixture settings, using $2 \times 2$ average-pooling before the classifier layer instead of the $4 \times 4$ average-pooling used by \citet{lopez2017gradient}.
Moreover, we present the dimension of the representation space of each layer of our architectures in Table~\ref{tab:layer-dimension}.

\begin{table}[h]
    \centering
    \caption{Dimension of the representation space of each layer.}
    \begin{tabular}{lp{3cm}p{8cm}}
        \toprule
        Network & Depth & Dimension of the representation space \\
        \midrule
        MLP & 3 layers & 784; 100; 100\\
        AlexNet & 5 layers & 48; 576; 512; 1,024; 2,048 \\
        LeNet-5 & 4 layers & 75; 500; 3,200; 800 \\
        ResNet-18 & 17 layers and 3 short-cut connections & 27; 180; 180; 180; 180; 180; 360; 20; 360; 360; 360; 720; 40; 720; 720; 720; 1,440; 80; 1,440; 1,440 \\
        \bottomrule
    \end{tabular}
    \label{tab:layer-dimension}
\end{table}

\subsection{Implementation Details}
\label{app:implementation-detail}

We use the official implementation of GPM~\citep{saha2021gradient}, HAT~\cite{serra2018overcoming}, and TRGP~\citep{lin2022trgp}.
We implement A-GEM and ER\_Res with the official implementation by \citet{chaudhry2018efficient} and implement EWC with the implementation by \citet{serra2018overcoming}.
For OGD~\citep{farajtabar2020orthogonal}, we use the implementation by \citet{bennani2020generalisation}.
Following \citet{saha2021gradient} and \citet{lin2022trgp}, we run all experiments five times on an established seed without fixing the cuda settings for a fair comparison.
Particularly, we use five random seeds on PMNIST where there is no diversity on a single seed. 
For CIFAR-100 Sup, we use five different orders provided by \citet{yoon2019scalable}.
Following \citet{saha2021gradient}, we report the experimental results of replay-base methods A-GEM and ER\_Res on the Mixture dataset with the same buffer size as GPM and ROGO, which is 8.98M in terms of the number of parameters for Resnet18 architecture.

On CIFAR-100 Split, MiniImageNet, and PMNIST, we follow the hyper-parameters utilized by \citet{saha2021gradient} and \citet{lin2022trgp}, including learning rate, batch size, and the threshold $\epsilon$.
On Mixture, as we adopt the same network architecture \citet{saha2021gradient} use on their 5-Dataset setting, we follow the provided learning rate and batch size as well.

As discussed in Section~\ref{sec:5}, the threshold $\zeta = \cos{\gamma}$ controls the criterion of the relaxing space.
For CIFAR100-Split and PMNIST, we use $\zeta=0.95$ for convolutional layers and $\zeta=0.9$ for fully connected layers.
For MiniImageNet and Mixture, we use the same $\zeta$ for all layers, $0.95$ and $0.9$ respectively.
The regularization weight $\beta$ is set as 5 for the ResNet18 architecture and 1 for others.
Moreover, we limit the max iteration times to 2 for CIFAR100-Split and MiniImageNet for efficiency.
Particularly, we run all the experiments on a single NVIDIA GeForce RTX 2080 Ti GPU.

\subsection{Metrics}
\label{app:metric}

Here we present the detailed definitions of the metrics evaluating the forward knowledge transfer.
$\Omega_{new}$~\citep{kemker2018measuring}.
Denote $b_i$ as the test accuracy of task $i$ at random initialization, FWT, first proposed by \citet{lopez2017gradient}, is defined as $\mathrm{FWT} = \frac{1}{T-1} \sum^T_{i=2} (A_{i-1,i}-b_i)$, evaluating the zero-shot performance of the initialization with respect to the observed tasks.
While $\Omega_{new}$, first proposed by \citet{kemker2018measuring}, is defined as $\Omega_{new} = \frac{1}{T-1} \sum^T_{i=2} (A_{i,i}-b_i)$, reflecting the test accuracy on new tasks based on the learnt knowledge.
As $b_i$ stays still across different approaches, we consider $\Omega_{new} = \frac{1}{T-1} \sum^T_{i=2} A_{i,i}$ for simplicity.
For this simplified $\Omega_{new}$, we have: $\Omega_{new} = \frac{T}{T-1} ACC - BWT - \frac{1}{T-1} A_{1,1}$, with the ACC and BWT defined in Section~\ref{sec:4-1}.

\section{Experimental Results}
\label{app:exp-results}

\subsection{Final Accuracy}
\label{app:final-acc}

We provide the test accuracy after learning each task on other benchmarks here.
As discussed in Section~\ref{sec:4}, our ROGO universally outperforms GPM over the task sequence on all benchmarks.

\begin{figure}[h!]
    \centering
    \includegraphics[width=\linewidth]{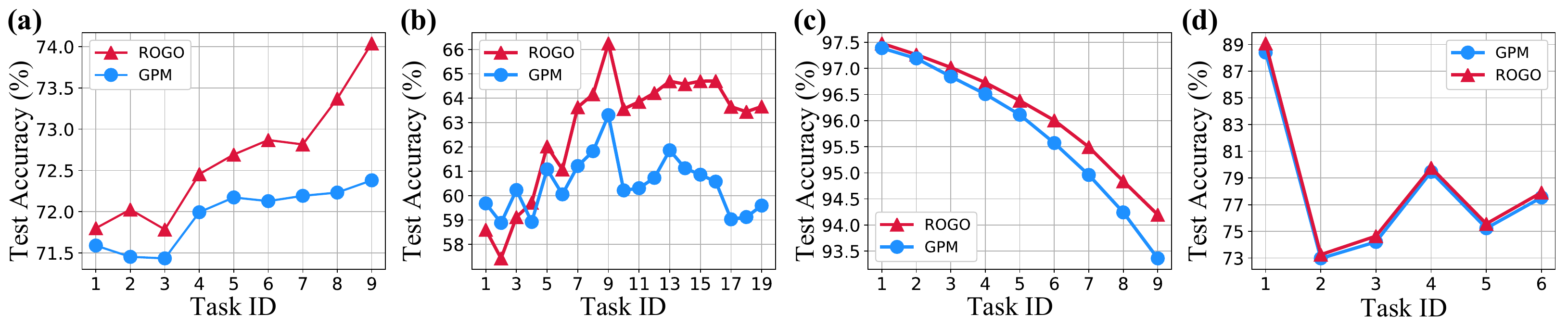}
    \caption{Average accuracy after learning each task on (a) CIFAR100-Split, (b) MiniImageNet, (c) PMNIST, and (d) Mixture.}
    \label{fig:final-acc-all}
\end{figure}

\subsection{Backward Knowledge Trasnfer}
\label{app:bwt}

As mentioned in Section~\ref{sec:4-1}, final accuracy (ACC), forgetting (BWT) and forward knowledge transfer ($\Omega_{new}$) are jointly considered to evaluate a continual learner.
According to Table~\ref{tab:main-bwt}, ROGO achieves the best forgetting on MiniImageNet and PMNIST datasets.
Despite our relaxing strategy focusing on facilitating the forward knowledge transfer, ROGO gains better final accuracy with comparable forgetting over all benchmarks.
Generally speaking, ROGO achieves superior performance than previous baselines with a fixed network capacity.

\begin{table*}[h]
\caption{Comparison of average accuracy and forgetting tested after learning all tasks. \emph{Multitask} is under the non-incremental setting. For each task, we mark the best and the second best performance in \textbf{bold} and \underline{underline} respectively. All results reported are averaged over 5 runs.}
\vspace{-5pt}
\label{tab:main-bwt}
\begin{center}
\scalebox{0.82}{
\begin{tabular}{lrrrrrrrr}
\toprule
\multirow{2}{*}{\bf Method}  & \multicolumn{2}{c}{\bf CIFAR-100 Split} & \multicolumn{2}{c}{\bf MiniImageNet} & \multicolumn{2}{c}{\bf PMNIST} & \multicolumn{2}{c}{\bf Mixture} \\
\cmidrule(lr){2-3} \cmidrule(lr){4-5} \cmidrule(lr){6-7} \cmidrule(lr){8-9}
& ACC (\%) & BWT (\%) & ACC (\%) & BWT (\%) & ACC (\%) & BWT (\%) & ACC (\%) & BWT (\%) \\
\midrule
Multitask & 79.58 $\pm$ 0.54 & - & 69.46 $\pm$ 0.62 & - & 96.70 $\pm$ 0.02 & - & 81.29 $\pm$ 0.23 & - \\
\midrule
A-GEM & 63.98 $\pm$ 1.22 & -16.30 $\pm$ 1.19 & 57.24 $\pm$ 0.72 & -10.95 $\pm$ 1.29 & 83.56 $\pm$ 0.16 & -14.57 $\pm$ 2.33 & 59.86 $\pm$ 1.01 & -29.37 $\pm$ 1.11 \\
ER\_Res & 71.73 $\pm$ 0.63 & -5.50 $\pm$ 0.76 & 58.94 $\pm$ 0.85 & -8.84 $\pm$ 0.56 & 87.24 $\pm$ 0.53 & -10.24 $\pm$ 1.58 & 75.07 $\pm$ 0.55 & -11.81 $\pm$ 0.82 \\
\midrule
EWC & 68.80 $\pm$ 0.88 & -2.31 $\pm$ 0.76 & 52.01 $\pm$ 2.53 & -12.14 $\pm$ 2.28 & 89.97 $\pm$ 0.57 & -3.58 $\pm$ 1.22 & 69.62 $\pm$ 2.69 & -6.00 $\pm$ 4.09 \\
HAT & 72.06 $\pm$ 0.50 & \bf -0.12 $\pm$ 0.42 & 59.78 $\pm$ 0.57 & -3.31 $\pm$ 0.05 & - & - & \underline{77.54 $\pm$ 0.18} & \bf -0.55 $\pm$ 0.28 \\
GPM & \underline{72.48 $\pm$ 0.40} & \underline{-0.72 $\pm$ 0.27} & \underline{60.41 $\pm$ 0.61} & \underline{-1.54 $\pm$ 0.34} & \underline{93.91 $\pm$ 0.16} & \underline{-3.30 $\pm$ 0.09} & 77.49 $\pm$ 0.68 & -4.70 $\pm$ 0.50 \\
\midrule
ROGO & \bf 74.04 $\pm$ 0.35 & -1.43 $\pm$ 0.43 & \bf 63.66 $\pm$ 1.24 & \bf -0.09 $\pm$ 0.94 & \bf 94.20 $\pm$ 0.11 & \bf -2.44 $\pm$ 0.13 & \bf 77.91 $\pm$ 0.45 & \underline{-4.55 $\pm$ 0.36} \\
\bottomrule
\end{tabular}
}
\end{center}
\vspace{-5pt}
\end{table*}

\subsection{Forward Knowledge Transfer}
\label{app:fwt}

We provide detailed forward knowledge transfer performance on all four benchmarks here.
First, we present the results of $\Omega_{new}$ and the detailed accuracy of each task after learning it in Table~\ref{tab:fwt-cifar100} to~\ref{tab:fwt-mixture}.
According to Table~\ref{tab:fwt-pmnist} and Table~\ref{tab:fwt-mixture}, ROGO achieves a similar forward knowledge transfer compared with GPM.
For other benchmarks, ROGO improves $\Omega_{new}$ by 2.7\% and 1.8\% on CIFAR-100 Split and MiniImageNet respectively, as shown in Table~\ref{tab:fwt-cifar100} and Table~\ref{tab:fwt-imagenet}.

\begin{table}[h!]
\caption{The accuracy tested on task $i$ after learning task $i$ and $\Omega_{new}$ on CIFAR-100 Split.}
\label{tab:fwt-cifar100}
\begin{center}
\scalebox{0.95}{
\begin{tabular}{lcccccccccc}
\toprule
Method & 1 & 2 & 3 & 4 & 5 & 6 & 7 & 8 & 9 & Avg ($\Omega_{new}$) \\
\midrule
GPM & 67.7 & 72.5 & 70.1 & 73.6 & 71.8 & 70.3 & 71.0 & 71.8 & 73.7 & 71.5 \\
ROGO & 71.7 & 74.8 & 73.6 & 76.9 & 75.8 & 74.7 & 74.0 & 75.9 & 79.3 & 75.2 \\
\bottomrule
\end{tabular}
}
\end{center}
\end{table}
\begin{table}[h!]

\caption{The accuracy tested on task $i$ after learning task $i$ and $\Omega_{new}$ on MiniImageNet.}
\label{tab:fwt-imagenet}
\begin{center}
\scalebox{0.65}{
\begin{tabular}{lcccccccccccccccccccc}
\toprule
Method & 1 & 2 & 3 & 4 & 5 & 6 & 7 & 8 & 9 & 10 & 11 & 12 & 13 & 14 & 15 & 16 & 17 & 18 & 19 & Avg ($\Omega_{new}$) \\
\midrule
GPM & 59.0 & 57.1 & 65.6 & 59.6 & 76.2 & 57.2 & 66.5 & 72.8 & 81.5 & 42.0 & 59.6 & 62.1 & 62.3 & 58.9 & 57.2 & 59.0 & 50.7 & 65.7 & 57.3 & 61.6 \\
ROGO & 57.1 & 55.1 & 66.2 & 61.3 & 79.8 & 61.4 & 66.6 & 73.5 & 83.6 & 42.8 & 65.1 & 63.4 & 67.1 & 64.6 & 61.6 & 61.6 & 50.0 & 68.2 & 60.8 & 63.7 \\
\bottomrule
\end{tabular}
}
\end{center}
\end{table}

\begin{table}[h!]

\caption{The accuracy tested on task $i$ after learning task $i$ and $\Omega_{new}$ on PMNIST.}
\label{tab:fwt-pmnist}
\begin{center}
\scalebox{0.9}{
\begin{tabular}{lcccccccccc}
\toprule
Method & 1 & 2 & 3 & 4 & 5 & 6 & 7 & 8 & 9 & Avg ($\Omega_{new}$) \\
\midrule
GPM & 97.5 & 97.4 & 97.0 & 96.8 & 96.5 & 96.2 & 96.2 & 95.8 & 95.1 & 96.5 \\
ROGO & 97.4 & 97.2 & 97.0 & 96.5 & 96.3 & 96.1 & 95.7 & 94.9 & 94.8 & 96.2 \\
\bottomrule
\end{tabular}
}
\end{center}
\end{table}

\begin{table}[h!]

\caption{The accuracy tested on task $i$ after learning task $i$ and $\Omega_{new}$ on Mixture.}
\label{tab:fwt-mixture}
\begin{center}
\scalebox{0.9}{
\begin{tabular}{lccccccc}
\toprule
Method & 1 & 2 & 3 & 4 & 5 & 6 & Avg ($\Omega_{new}$) \\
\midrule
GPM & 99.0 & 43.7 & 87.3 & 99.1 & 69.9 & 93.4 & 82.1 \\
ROGO & 99.1 & 42.9 & 87.6 & 99.1 & 70.6 & 93.5 & 82.2 \\
\bottomrule
\end{tabular}
}
\end{center}
\end{table}

Moreover, we provide the result of FWT (using the definition in \citep{lopez2017gradient}) on all benchmarks in Table~\ref{tab:fwt-app}.
According to Table~\ref{tab:fwt-app}, although our method facilitates the forward knowledge transfer reflected by $\Omega_{new}$, ROGO achieves better FWT than GPM on all three task-incremental benchmarks, namely better zero-shot performance.

\begin{table}[h!]
\caption{Comparison of forward knowledge transfer between GPM and ROGO, evaluated by FWT.}
\label{tab:fwt-app}
\begin{center}
\scalebox{0.85}{
\begin{tabular}{lrrrr}
\toprule
Methods & \bf CIFAR-100 Split & \bf MiniImageNet & \bf PMNIST & \bf Mixture \\
\midrule
GPM & -0.65 $\pm$ 0.18 & -0.26 $\pm$ 0.88 & \bf +0.66 $\pm$ 1.17 & -0.52 $\pm$ 1.49 \\
ROGO & \bf +0.20 $\pm$ 0.36 & \bf +0.30 $\pm$ \bf 0.29 & -0.63 $\pm$ 0.97 & \bf +0.13 $\pm$ \bf 0.77 \\
\bottomrule
\end{tabular}
}
\end{center}
\end{table}

\subsection{Accuracy Evolution}
\label{app:acc-evo}

Here we present the accuracy tested on three randomly selected tasks immediately after learning them.
We select the 2nd, 4th, and 6th tasks for all benchmarks.
Generally, ROGO outperforms GPM on selected tasks over the sequence.
We further notice that the improvement is more significant on later tasks owing to a larger relaxing space, as discussed in Section~\ref{sec:5}.

\begin{figure}[h!]
    \centering
    \includegraphics[width=\linewidth]{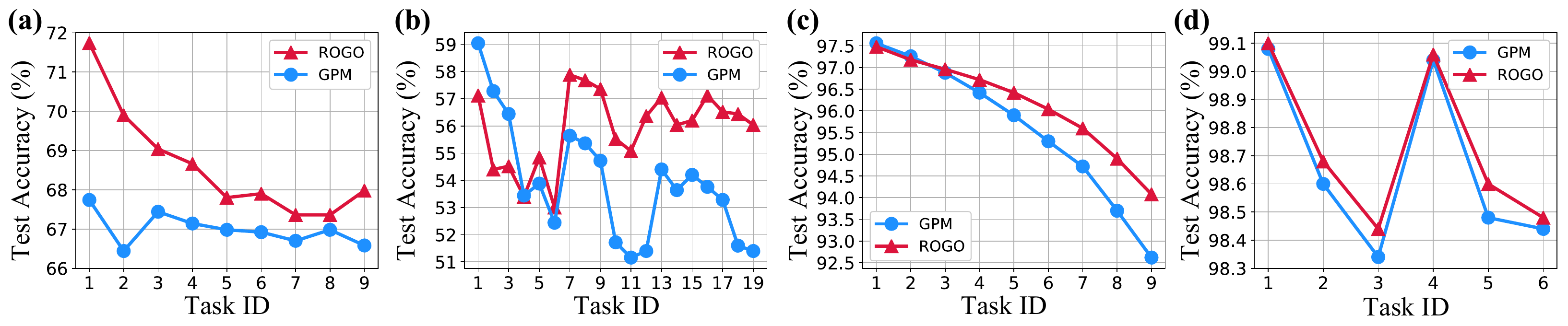}
    \caption{Accuracy evolution of the 2nd task on (a) CIFAR-100 Split, (b) MiniImageNet, (c) PMNIST, and (d) Mixture.}
    \label{fig:acc-evo-2}
\end{figure}

\begin{figure}[h!]
    \centering
    \includegraphics[width=\linewidth]{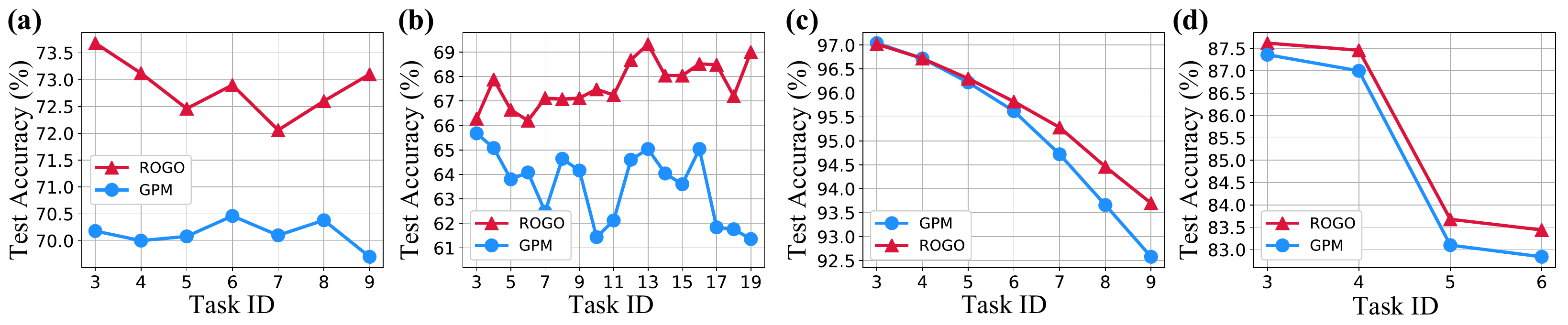}
    \caption{Accuracy evolution of the 4th task on (a) CIFAR-100 Split, (b) MiniImageNet, (c) PMNIST, and (d) Mixture.}
    \label{fig:acc-evo-4}
\end{figure}

\begin{figure}[h!]
    \centering
    \includegraphics[width=\linewidth]{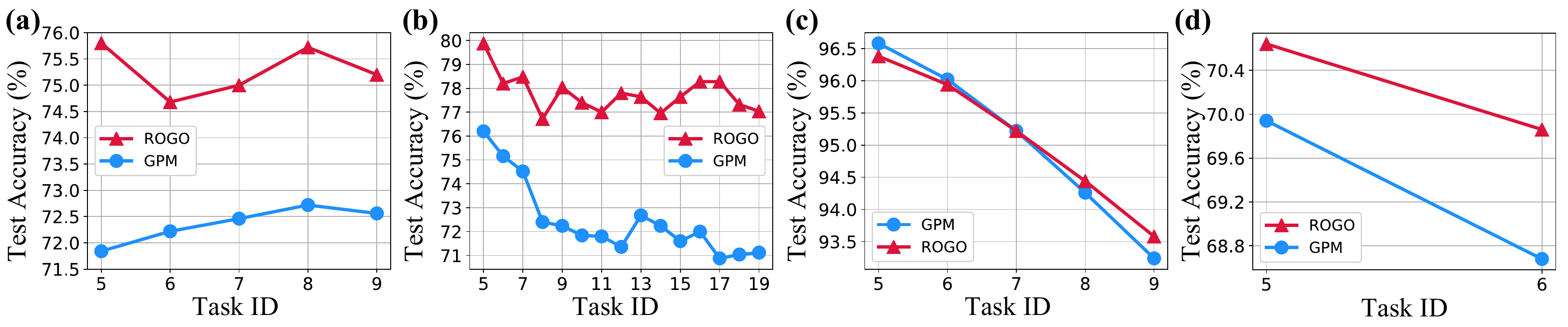}
    \caption{Accuracy evolution of the 6th task on (a) CIFAR-100 Split, (b) MiniImageNet, (c) PMNIST, and (d) Mixture.}
    \label{fig:acc-evo-6}
\end{figure}

\subsection{Time Consumption}
\label{app:time-consume}

We report the time consumption of ROGO on two benchmarks compared with relative baselines.
TRGP and ROGO are both evaluated on a single NVIDIA GeForce RTX 2080 Ti GPU and we report the results according to~\citep{lin2022trgp}.
As discussed in Section~\ref{sec:5}, ROGO takes acceptable extra time compared with GPM on both datasets.
For both dataset, ROGO tasks similar time as TRGP, which is similar to EWC and much less than A-GEM and OWM.

\begin{table}[h!]
\caption{Time comparison evaluated on two benchmarks. We use the results reported in~\citep{lin2022trgp} and the time is normalized with respect to GPM.}
\label{tab:time-comparison}
\begin{center}
\scalebox{0.9}{
\begin{tabular}{lcccccccc}
\toprule
\multirow{2}{*}{Datasets} & \multicolumn{8}{c}{Methods} \\
\cmidrule(lr){2-9}
& OWM & EWC & HAT & A-GEM & ER\_Res & GPM & TRGP & ROGO \\
\midrule
CIFAR-100 & 2.41 & 1.76 & 1.62 & 3.48 & 1.49 & 1.00 & 1.65 & 1.62 \\
MiniImageNet & - & 1.22 & 0.91 & 1.79 & 0.82 & 1.00 & 1.34 & 1.43 \\
\bottomrule
\end{tabular}
}
\end{center}
\end{table}

\subsection{Memory Usage}
\label{app:memory-usage}

\begin{figure}[h!]
    \centering
    \includegraphics[width=\linewidth]{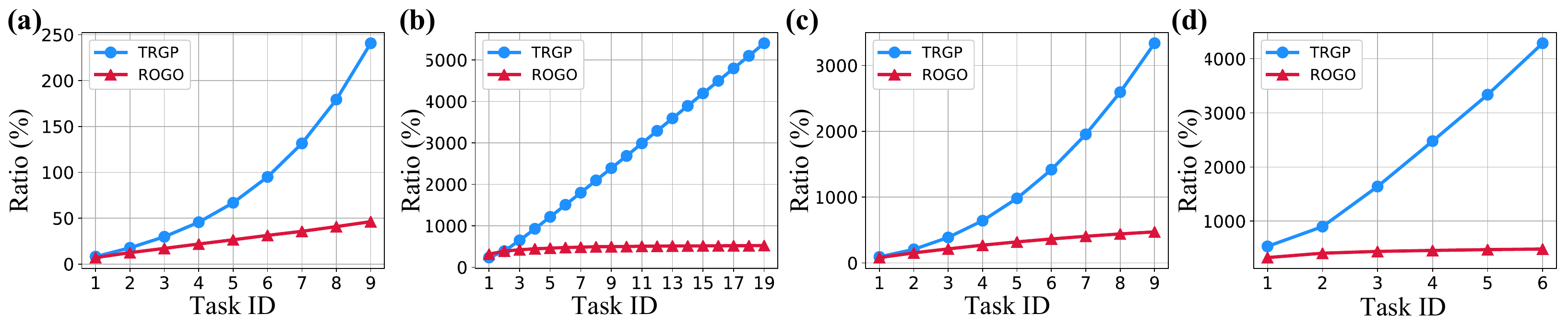}
    \caption{Ratio of the amount of extra parameters concerning the amount of the parameters of the initial network architecture on (a) CIFAR100-Split, (b) MiniImageNet, (c) PMNIST, and (d) Mixture.}
    \label{fig:extra-memory}
\end{figure}

We provide a comparison between TRGP and ROGO on the ratio of the amount of extra parameters concerning the amount of the parameters of the initial network architecture.
According to Figure~\ref{fig:extra-memory}, TRGP requires at least 200\% of the number of extra parameters after learning all tasks on all four benchmarks, while ROGO only stores the representation of the frozen space, which can further be released in the inference phase.

\subsection{Other Results}
\label{app:others}

Here we provide detailed results including the forgetting performance compared with TRGP in Table~\ref{tab:trgp-compare-exp-bwt} and~\ref{tab:trgp-compare-non-exp-bwt}.
Similar to the discussion in Section~\ref{sec:4-3}, our ROGO framework obtains superior accuracy performance with comparable forgetting under or without the constraint of a fixed network capacity.

\begin{table*}[h!]
    \centering
    \caption{Comparison of average accuracy and forgetting with TRGP under an expansion setting. The percentages indicate the ratios of the rank of the relaxing space with respect to the frozen space. For each task, we mark the best and the second best performance in \textbf{bold} and \underline{underline}.}
    \label{tab:trgp-compare-exp-bwt}
    \scalebox{0.82}{
    \begin{tabular}{lrrrrrrrr}
        \toprule
        \multirow{3}{*}{Methods} & \multicolumn{2}{c}{\multirow{2}{*}{TRGP}} & \multicolumn{6}{c}{ROGO-Exp} \\
         & & & \multicolumn{2}{c}{50\%} & \multicolumn{2}{c}{80\%} & \multicolumn{2}{c}{T\%} \\
        \cmidrule(lr){2-3} \cmidrule(lr){4-5} \cmidrule(lr){6-7} \cmidrule(lr){8-9}
        & ACC (\%) & BWT (\%) & ACC (\%) & BWT (\%) & ACC (\%) & BWT (\%) & ACC (\%) & BWT (\%) \\
        \midrule
        CIFAR & 74.46 $\pm$ 0.22 & \bf -0.42 $\pm$ 0.20 & 74.90 $\pm$ 0.37 & -0.99 $\pm$ 0.27 & \underline{74.97 $\pm$ 0.30} & -0.68 $\pm$ 0.39 & \bf 75.34 $\pm$ 0.61 & \underline{-0.63 $\pm$ 0.57} \\
        PMNIST & 96.34 $\pm$ 0.11 & -0.58 $\pm$ 0.10  & 96.44 $\pm$ 0.20 & -0.75 $\pm$ 0.13 & \underline{96.76 $\pm$ 0.15} & \underline{-0.46 $\pm$ 0.08} & \bf 97.01 $\pm$ 0.08 & \bf -0.31 $\pm$ 0.05 \\
        MiniImageNet & 61.78 $\pm$ 1.94 & -1.01 $\pm$ 0.58 & \bf 63.46 $\pm$ 0.85 & \bf 0.50 $\pm$ 0.39 & 62.57 $\pm$ 1.32 & 0.42 $\pm$ 0.57 & \underline{62.78 $\pm$ 1.00} & \underline{0.45 $\pm$ 0.35} \\
        Mixture & \underline{83.54 $\pm$ 1.15} & -0.80 $\pm$ 1.10 & 82.45 $\pm$ 0.49 & \underline{-0.74 $\pm$ 0.16} & 83.33 $\pm$ 0.31 & -0.98 $\pm$ 0.23 & \bf 83.62 $\pm$ 0.26 & \bf -0.33 $\pm$ 0.14 \\
        \bottomrule
    \end{tabular}
    }
\end{table*}

\begin{table*}[h!]
    \centering
    \caption{Comparison of average accuracy and forgetting with TRGP within a fixed network capacity. We modify TRGP as TRGP-Reg with similar regularization terms. $\beta$ indicates the regularization weight. For each task, we mark the best and the second best performance in \textbf{bold} and \underline{underline}.}
    \label{tab:trgp-compare-non-exp-bwt}
    \scalebox{0.8}{
    \begin{tabular}{lrrrrrrrr}
        \toprule
        \multirow{3}{*}{Methods} & \multicolumn{6}{c}{TRGP-Reg} & \multicolumn{2}{c}{\multirow{2}{*}{ROGO}} \\
         & \multicolumn{2}{c}{$\beta=1$} & \multicolumn{2}{c}{$\beta=5$} & \multicolumn{2}{c}{$\beta=50$} & & \\
        \cmidrule(lr){2-3} \cmidrule(lr){4-5} \cmidrule(lr){6-7} \cmidrule(lr){8-9}
        & ACC (\%) & $\Omega_{new}$ (\%) & ACC (\%) & $\Omega_{new}$ (\%) & ACC (\%) & $\Omega_{new}$ (\%) & ACC (\%) & $\Omega_{new}$ (\%) \\
        \midrule
        CIFAR & 72.01 $\pm$ 0.30 & -1.63 $\pm$ 0.20 & 71.96 $\pm$ 0.40 & \underline{-1.09 $\pm$ 0.37} & \underline{72.49 $\pm$ 0.09} & \bf -0.69 $\pm$ 0.24 & \bf 74.04 $\pm$ 0.35 & -1.43 $\pm$ 0.43 \\
        PMNIST & \underline{76.57 $\pm$ 2.69} & \underline{-20.69 $\pm$ 2.94} & 75.50 $\pm$ 3.96 & -22.41 $\pm$ 4.24 & 76.56 $
        \pm$ 3.83 & -21.63 $\pm$ 4.13 & \bf 94.20 $\pm$ 0.11 & \bf -0.09 $\pm$ 0.94 \\
        MiniImageNet & 55.81 $\pm$ 2.43 & -3.69 $\pm$ 2.19 & \underline{58.81 $\pm$ 2.24} & \bf -2.85 $\pm$ 1.54 & 22.69 $\pm$ 0.39 & 0.01 $\pm$ 0.07 & \bf 63.66 $\pm$ 1.24 & \underline{-2.44 $\pm$ 0.13} \\
        Mixture & 73.31 $\pm$ 1.05 & -11.05 $\pm$ 1.43 & \underline{74.71 $\pm$ 0.85} & -9.33 $\pm$ 1.05 & 17.36 $\pm$ 0.86 & \bf -0.01 $\pm$ 0.03 & \bf 77.91 $\pm$ 0.45 & \underline{-4.55 $\pm$ 0.36} \\
        \bottomrule
    \end{tabular}
    }
\end{table*}

\begin{figure}[h!]
    \centering
    \includegraphics[width=\linewidth]{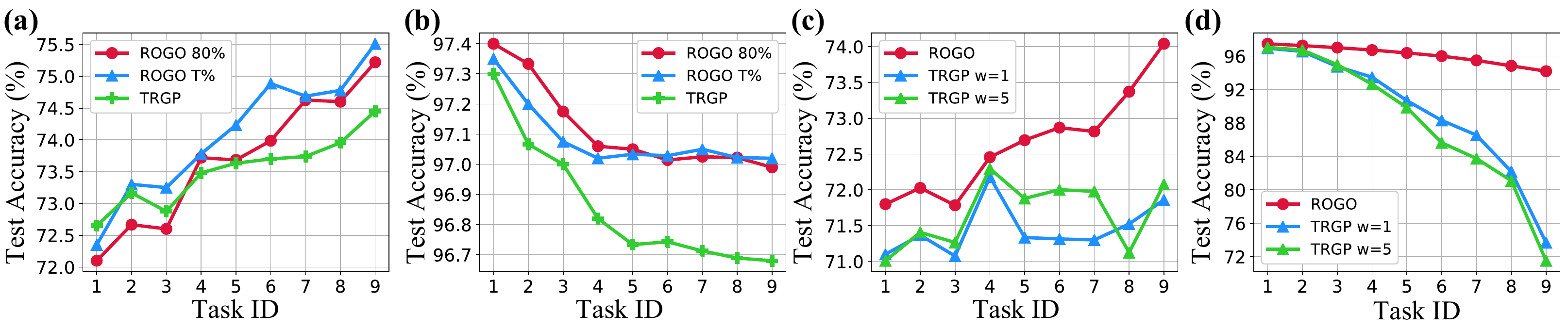}
    \caption{Test accuracy after learning each task under an expansion setting on (a) CIFAR-100 Split and (b) PMNIST, and within a fixed network capacity on (c) CIFAR-100 Split and (d) PMNIST.}
    \label{fig:trgp-exp-reg}
\end{figure}

Moreover, we illustrate the test accuracy over the task sequence in Figure~\ref{fig:trgp-exp-reg}.
As shown in Figure 9-(a) and Figure 9-(b), our ROGO-Exp dominates TRGP on both benchmarks relaxing either 80\% or T\% of the frozen space under an expansion setting.
We further compare ROGO with TRGP modified with the same regularization terms.
As shown in Figure 9-(c) and Figure 9-(d), the performance of TRGP drops significantly constrained in a fixed network capacity on both benchmarks.

Following GPM~\citep{saha2021gradient}, we conduct experiments on the CIFAR-100 Sup dataset as well.
The implementation details are included in Appendix~\ref{app:implementation-detail} as well.
Here we report the accuracy performance in Table~\ref{tab:cifar-sup}.
Note that the selected baselines for this setting all require extra parameters other than GPM and we directly use the results of the baselines reported in GPM and TRGP.
In accordance with the results on the other four benchmarks discussed in Section~\ref{sec:4-2}, ROGO outperforms GPM and other related baselines by a significant margin.
Particularly, we notice ROGO directly obtains superior performance than TRGP within a fixed network capacity under this setting.

\begin{table}[h]
\caption{Results of ACC (\%) on CIFAR-100 Sup setting. \emph{STL} is under a non-incremental setting. All baselines require extra network capacity except GPM. We also report the standard deviation result of ROGO for comparison.}
\vspace{-5pt}
\label{tab:cifar-sup}
\begin{center}
\scalebox{0.8}{
\begin{tabular}{lcccccccc}
\toprule
\multirow{2}{*}{Metric} & \multicolumn{8}{c}{Methods} \\
\cmidrule(lr){2-9}
& STL* & PNN & DEN & RCL & APD & GPM & TRGP & ROGO \\
\midrule
ACC (\%) & 61.00 & 50.76 & 51.10 & 51.99 & 56.81 & 57.72 & 58.25 & \bf 58.74 $\pm$ 0.60 \\
\bottomrule
\end{tabular}
}
\end{center}
\vspace{-15pt}
\end{table}

\section{Algorithm}
\label{app:algo}

We present the pseudo-code of our searching strategy here.

\begin{algorithm} 
    \renewcommand{\algorithmicrequire}{\textbf{Input:}}
	\renewcommand{\algorithmicensure}{\textbf{Output:}}
	\caption{Relaxing Space Searching}
	\label{alg:RSS}
	\begin{algorithmic}[1]
	    \REQUIRE gradient $\{g^l_t\}_{l=1}^{L}$, frozen space $ \{U_{t-1}^l\}_{l=1}^{L}$ and thresholds $\{\epsilon^l_{th}, \gamma^l_t\}_{l=1}^{L}$
		\ENSURE relaxing space $\{V_t^l\}_{l=1}^{L}$
		\FOR {$l \in 1,...,L$}
		    \STATE Construct the significant representation space $R_{g,t}^l$ from gradients $g^l_t$ with top-$k$ eigenvectors.
		    \STATE $V_t^l \leftarrow \varnothing$
		    \REPEAT
		    \STATE $d \leftarrow \underset{d \in U^{l,c}_{t-1}}{\mathop{\arg\min}} \mathit{\Theta} ( d, R_{g,t}^l )$
		    \IF{$\mathit{\Theta} ( d, R_{g,t}^l ) \leq \gamma^l_t$}
                \STATE $V_t^l \leftarrow V_t^l \cup d$ \STATE $U^{l,c}_{t-1} \leftarrow U^l_{t-1} \backslash V^l_t$
		    \ENDIF
		    \UNTIL $\mathit{\Theta} ( d, R_{g,t}^l ) > \gamma^l_t$
        \ENDFOR
    \end{algorithmic}
    \label{searching}
\end{algorithm}

\end{document}